\documentclass{article}

\usepackage[nonatbib, final]{neurips_glfrontiers_2023}

\usepackage[utf8]{inputenc} 
\usepackage[T1]{fontenc}    
\usepackage{thmtools, hyperref}       
\usepackage{url}            
\usepackage{booktabs}       
\usepackage{amsfonts}       
\usepackage{nicefrac}       
\usepackage{microtype}      
\usepackage{xcolor}         
\usepackage[pdftex]{graphicx}
\usepackage{epstopdf}
\usepackage{color, colortbl}
\usepackage{multirow, makecell}
\usepackage{rotating}

\usepackage{subcaption}
\usepackage{wrapfig}
\newtheorem{theorem}{Theorem}
\usepackage{algorithm}
\usepackage[noend]{algpseudocode}
\usepackage{amsmath}

\usepackage{color}
\definecolor{Gray}{gray}{0.9}
\usepackage{bm}
\usepackage{enumitem}
\makeatletter
\def\algbackskip{\hskip-\ALG@thistlm}
\makeatother

\DeclareMathOperator*{\argmax}{argmax}
\DeclareMathOperator*{\argmin}{argmin}
\usepackage{bbm}

\newtheorem{lemma}{Lemma}
\newtheorem{assumption}{Assumption}

\title{On the Temperature of Bayesian Graph Neural Networks for Conformal Prediction}

\author{%
  Seohyeon ~Cha \\
  KAIST \\
  \texttt{kaitjgus@kaist.ac.kr} \\
  \And
  Honggu ~Kang \\
  KAIST\\
  \texttt{khg13@kaist.ac.kr} \\
  \And
  Joonhyuk ~Kang \\
  KAIST \\
  \texttt{jkang@kaist.ac.kr} \\
}

\begin{document}

\maketitle

\begin{abstract}
Accurate uncertainty quantification in graph neural networks (GNNs) is essential, especially in high-stakes domains where GNNs are frequently employed. Conformal prediction (CP) offers a promising framework for quantifying uncertainty by providing \textit{valid} prediction sets for any black-box model. CP ensures formal probabilistic guarantees that a prediction set contains a true label with a desired probability. However, the size of prediction sets, known as \textit{inefficiency}, is influenced by the underlying model and data generating process. On the other hand, Bayesian learning also provides a credible region based on the estimated posterior distribution, but this region is \textit{well-calibrated} only when the model is correctly specified. Building on a recent work that introduced a scaling parameter for constructing valid credible regions from posterior estimate, our study explores the advantages of incorporating a temperature parameter into Bayesian GNNs within CP framework. We empirically demonstrate the existence of temperatures that result in more efficient prediction sets. Furthermore, we conduct an analysis to identify the factors contributing to inefficiency and offer valuable insights into the relationship between CP performance and model calibration. 
\end{abstract}

\section{Introduction}
Graph neural networks (GNNs) have demonstrated their impressive abilities to learn from graph-structured data in a wide range of domains, including social sciences, chemistry, knowledge graphs, and recommendation systems \cite{kipf2016semi, teixeira2019graph, hamilton2017inductive, xu2018powerful, dwivedi2020benchmarking}. Meanwhile, with the increasing applications of GNNs in safety-critical tasks, there is a growing demand for trustworthy and reliable uncertainty estimates from GNN outputs 
\cite{hsu2022makes, wang2021confident}. However, many existing uncertainty quantification methods are not applicable to GNNs, since they rely on independent and identically distributed (i.i.d.) data assumption, which is violated in graph-structured data \cite{abdar2021review, hullermeier2021aleatoric}. Recent studies have addressed uncertainty in graph-related tasks, but these approaches introduce additional model architecture for density estimation or knowledge distillation.
\cite{stadler2021graph, zhao2020uncertainty}.

Conformal prediction (CP) \cite{vovk2005algorithmic} is a promising framework for obtaining uncertainty estimates under the only assumption of data exchangeability\footnote{The data exchangeability assumes that every permutation of data samples have the same probability, which is a weaker assumption than i.i.d assumption.}. CP constructs prediction sets that are guaranteed to contain a ground-truth with a desired coverage level. 
Formally, consider a training dataset $(x_i, y_i)\in \mathcal{X}\times\mathcal{Y}, \; i \in \{1, \cdots, n\}$, and a test point $(x_{n+1}, y_{n+1})$ drawn exchangeably from an underlying distribution $P$. Then, for a given a pre-defined coverage $1-\alpha$, CP generates a prediction set $\mathcal{C}_{n, \alpha} (x_{n+1})\subseteq \mathcal{Y}$ for a test input $x_{n+1}$ that satisfies
\begin{equation}
\mathbb{P}[y_{n+1} \in \mathcal{C}_{n, \alpha}(x_{n+1})] \ge 1-\alpha 
\label{eq1}
\end{equation}
where the probability is taken over $n+1$ data samples $\{(x_i, y_i)\}_{i=1}^{n+1}$. A prediction set satisfying the coverage condition in the equation \eqref{eq1} is said to be \textit{valid}. While CP guarantees a valid prediction for any classifier, the size of a prediction set, called \textit{inefficiency}, is largely affected by the underlying classifier and the data generating process \cite{dhillon2023expected, yang2021finite}. 

Meanwhile, Bayesian learning \cite{blundell2015weight, gal2016dropout, wilson2020bayesian} (see also \cite[Ch. 12]{simeone2022machine} for an overview) is a traditional way for building a predictive interval. In contrast to frequentist learning, which aims to find an optimal point of model parameters, Bayesian learning produces a posterior estimate for model parameters. Consequently, it can naturally generate credible regions by simply taking $(1-\alpha) \cdot 100\;$\% confidence region of the posterior distribution. However, the region can only be \textit{valid} without model misspecification, which means that prior, likelihood, and computational capability should be correctly specified in posterior inference \cite{masegosa2020learning, wilson2020bayesian}. To address the issue of model misspecification when constructing a posterior credible region, a recent study has introduced a scaling parameter that controls the spread of posterior \cite{syring2019calibrating} (see also \cite{knoblauch2019generalized} for related discussion on \emph{generalized Bayesian learning}). Their findings demonstrate that the control of the scaling parameter can yield a credible region with an appropriate size but provides approximate validity, whereas CP offers a rigorous guarantee of validity.

Building upon insights from previous works, we study the benefits of Bayesian GNNs in CP framework to obtain valid and efficient uncertainty estimates. While recent CP approaches for GNNs have focused on ensuring validity and reducing inefficiency, these methods are applicable to specific tasks \cite{clarkson2023distribution, zargarbashi2023conformal} or require correction dataset for additional GNN training \cite{ huang2023uncertainty}. In our work, we propose the use of a temperature parameter in Bayesian GNNs to allow for flexible control of inefficiency within CP framework. We show that Bayesian GNNs improve the performance of CP, especially the inefficiency, compared to frequentist GNNs while preserving the validity. Furthermore, our experiments on both node classification and graph classification tasks demonstrate the existence of temperatures of Bayesian GNNs that lead to more efficient prediction sets. While previous studies have explored the temperature parameter in Bayesian learning mainly for calibration purposes \cite{syring2019calibrating, zecchin2022robust}, our study investigates the impact of temperatures on CP performance. Moreover, we provide an analysis that explores the connection between the CP performance and model calibration, providing valuable insights on the varying inefficiency depending on the temperature. 

\begin{figure}[!t]
  \begin{center}
  \includegraphics[width=1\linewidth]{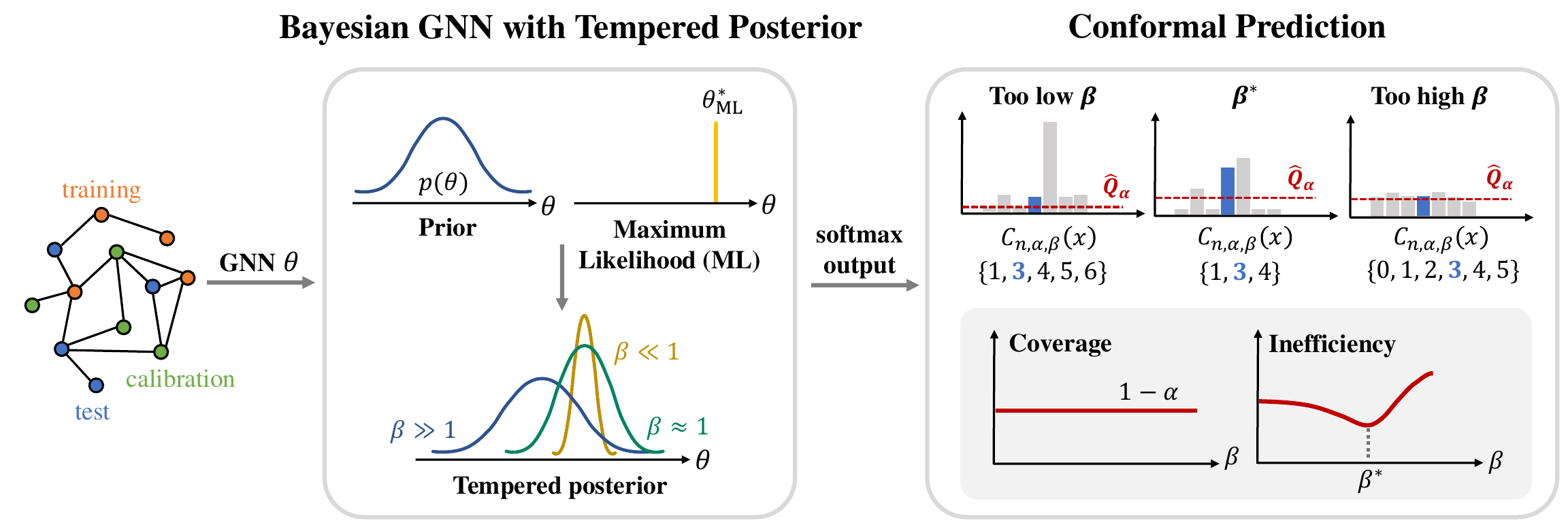}
  \end{center}
  \caption{\textbf{Conformal prediction (CP) of Bayesian GNNs with tempered posteriors}. In Bayesian GNNs, the temperature parameter $\beta$ controls the spread of the posterior over model parameters $\theta$. Once we train Bayesian GNNs with a specific temperature, CP generates prediction sets by comparing softmax output of a test sample to a threshold obtained from calibration set. In the right box of this figure, the blue bar represents the true label, and CP guarantees that prediction sets contain the true label with a pre-defined coverage. We show that Bayesian GNNs with an appropriate temperature produce more efficient prediction sets, reducing inefficiency, while keeping the desired coverage.}
  \label{fig:framework}
\end{figure}

The main contributions of this study are summarized as follows:
\begin{itemize}[topsep=0pt] 
    \item We show that Bayesian GNNs outperforms in CP compared to frequentist GNNs.
    \item Our experiments on graph-structured data demonstrate the existence of temperatures that result in better efficiency\footnote{Throughout the paper, we use the terms "lower inefficiency" and "better efficiency" interchangeably.}.
    \item We analyze the connection between the inefficiency and model calibration to verify the factors that contribute to the inefficiency. 
\end{itemize}

\section{Background}
\subsection{Conformal Prediction}

Conformal prediction (CP) \cite{vovk2005algorithmic} aims to construct a set of candidate labels that is likely to contain the true label. Formally, given a test example $x \in \mathcal{X}$ and for every candidate label $y \in \mathcal{Y}$, CP either rejects or accepts the candidate pair $(x, y)$ to be included in a prediction set. This decision is done based on a statistic, called  \textit{non-conformity score} $s((x, y) | \theta_\mathcal{D})$, a real-valued function that takes input-ouput pair $(x, y)$ as an input given a pre-trained model parameter $\theta_\mathcal{D}$. The score $s((x, y) |\theta_\mathcal{D})$ quantifies the disagreement between data sample $x$ and a candidate label $y$. One typical example of non-conformity score broadly used in classification task is negative log-likelihood ($-\log p_\theta(y|x)$).

In this work, we use one most common type of CP, called \textit{split conformal prediction (SCP)} \cite{vovk2005algorithmic}, which splits available dataset into training and calibration sets. Given a trained model on training set $\mathcal{D}$, SCP creates a prediction set $\mathcal{C}_{n, \alpha}(x)$ for a test input $x$
using the calibration set $\{(x_i, y_i)\}_{i=1}^n$. Specifically, we first compute the non-conformity scores of calibration samples as $\{s((x_i, y_i)|\theta_\mathcal{D})\}_{i=1}^n$. Then, the $(1-\alpha)$-quantile of the set $\{s((x_1, y_1)|\theta_\mathcal{D}), \dots, s((x_n, y_n)|\theta_\mathcal{D})\}\cup\{\infty\}$ becomes the threshold that determines whether to accept or reject candidate label $y \in \mathcal{Y}$ to be contained in the set.  The threshold can be obtained by selecting $\lceil (1-\alpha)(n+1)\rceil$-th smallest value from the set $\{s((x_1, y_1)|\theta_\mathcal{D}), \dots, s((x_n, y_n)|\theta_\mathcal{D})\}\cup\{\infty\}$, denoted as $Q_{\alpha}(\{s((x_i, y_i)|\theta_\mathcal{D})\}_{i=1}^n)$. Consequently, the corresponding prediction set for a test sample $x$ is defined as 
\begin{equation}
    \mathcal{C}_{n, \alpha}(x) =\left\{y\in\mathcal{Y}: s((x, y)|\theta_\mathcal{D}) \le Q_{\alpha}(\{s((x_i, y_i)|\theta_\mathcal{D})\}_{i=1}^n)\right\}.
    \label{eq:CPset} 
\end{equation}

\begin{theorem}
    Given that calibration set $\{(x_i, y_i)\}_{i=1}^n$ and a test data point $(x,y)$ are exchangeable random variables, for any coverage level $1-\alpha$, for any trained model $\theta_\mathcal{D}$, and for any nonconformity score $s(\cdot|\theta_\mathcal{D})$, a conformal prediction set $\mathcal{C}_{n, \alpha}(x)$ defined in the equation \eqref{eq:CPset} satisfies the equation \eqref{eq1} \normalfont{\cite{vovk2005algorithmic}}.
    \label{thm1: cov}
\end{theorem}
\autoref{thm1: cov} provides a marginal coverage guarantee for all $x$ on average. Roughly speaking, the SCP can induce any classifier to produce prediction sets guaranteed to satisfy the desired coverage level. This implies that, on average, these sets have a pre-defined probability of containing the true label. 

\subsection{Bayesian Learning}

We introduce two learning paradigms, \textit{frequentist learning} and \textit{Bayesian learning}, to find a classifier which predicts an output $y \in \mathcal{Y}$ for a given input $x \in \mathcal{X}$ . A conventional learning that finds an optimal model parameter $\theta$ and the corresponding single classifier $p(y|x, \theta)$ given training set $\mathcal{D} = \{x_i, y_i\}_{i=1}^n$, is called \textit{frequentist learning} \cite{murphy2012machine}. In general, it optimizes the model parameter by minimizing the training loss as 
\begin{equation}
    \theta^* = \argmin_\theta L_\mathcal{D} (\theta) =\sum_{(x, y) \in \mathcal{D}} -\log {p(y|x, \theta)}. 
    \label{eq:freq}
\end{equation}
Meanwhile, \textit{Bayesian learning} \cite{murphy2012machine, blundell2015weight, simeone2022machine} updates prior beliefs about model parameter $\theta$ that fits the dataset $\mathcal{D}$ while accounting for a prior belief $p(\theta)$ about the values of $\theta$. Instead of providing a single point estimate for $\theta$, it infers a posterior distribution over possible values of parameters $\theta$ based on Bayes' rule by incorporating likelihood from data and prior knowledge on $\theta$. Hence, it is readily capable of capturing and quantifying the uncertainty about model parameter $\theta$. Concretely, the Bayesian posterior can be obtained by minimizing the \textit{free energy criterion} \cite{knoblauch2022optimization, jose2021free},
\begin{equation}
   q_\beta^*(\theta|\mathcal{D}) = \argmin_{q} \mathbb{E}_{q(\theta)}[L_\mathcal{D}(\theta)] + \beta \cdot \text{KL}[q(\theta)||p(\theta)],
    \label{free energy} 
\end{equation}
where $\text{KL}[q(\theta)||p(\theta)]$ is the Kullback-Leibler (KL) divergence between two distributions $q$ and $p$, and $\beta > 0$ is the \textit{temperature} parameter which creates a tempered posterior $q_\beta^*(\theta|\mathcal{D})$. 

One advantage of posterior inference in Bayesian learning is that it can naturally produce credible regions from the posterior. For example, we simply can take 90\;\% confidence region of posterior distribution if we want to get a \textit{well-calibrated} (or \textit{valid}) region which satisfies the pre-defined coverage probability of 90\;\% to contain a target output in the region, as in the equation \eqref{eq1}. However, it fails in common when the model is \textit{misspecified} \cite{grunwald2017inconsistency, masegosa2020learning, knoblauch2019generalized}. Model misspecification occurs from three assumptions in posterior inference, which are prior belief, likelihood, and computational power required to perform inference on intractable posterior \cite{knoblauch2019generalized}. A model might converge to an incorrect solution if a true model is not included in the prior hypothesis space, or even if the space contains the true solution, a model may struggle to find a good solution unless it benefits from reasonable inductive biases from the likelihood \cite{wilson2020bayesian}. Hence, if there is model misspecification, the resulting posterior can be poorly calibrated and creates undesirable uncertainty quantification. (refer to Fig. 3 in \cite{knoblauch2019generalized}). There has been a study that introduces a scaling parameter to address the model misspecification when constructing a posterior credible region \cite{syring2019calibrating}. The scaling parameter controls the spread of the posterior distribution for the credible region to be \textit{efficient} in terms of its size and approximately \textit{well-calibrated}. Note that our work differs from previous studies in that we investigate how the temperature parameter $\beta$ affects the size of prediction sets obtained by CP when applied on top of Bayesian learning. 


\section{Conformal Prediction for Bayesian GNNs}

\vspace{-0.1cm}
\subsection{Bayesian GNNs with Tempered Posterior}
We begin with the explanation on how to implement Bayesian graph convolutional networks (GCNs). We use Graph DropConnect (GDC), a stochastic regularization technique with adaptive connection sampling \cite{hasanzadeh2020bayesian}. The adaptive connection sampling is a generalized technique that encompasses a range of other regularization and sampling techniques, including DropOut \cite{srivastava2014dropout}, DropEdge \cite{rong2019dropedge}, and Node Sampling \cite{chen2018fastgcn}. It operates by applying binary random masks to the adjacency matrix, randomly masking edges, nodes, and channels. They show that connection sampling in GDC can be interpreted as Bayesian extensions of GNNs by transferring randomness from sampling to the model parameter space. Consequently, by implementing GDC within GNNs, we are able to get an appropriate posterior estimate for GNN model parameters and further adjust its temperature. 

In detail, for every edge $e = (v, u) \in \mathcal{E}$, we generate a random mask vector denoted as $\mathbf{z}_{v,u}^{(l)} \in \{0,1\}^{1\times f_l}$ by sampling from Bernoulli distribution with a drop rate $\pi_l$, denoted as $\text{Bern}(\pi_l)$. Here, $f_l$ represents the number of features in the $l$-th layer of GNN. We define the degree of node $v$ as $c_v$ and its neighborhoods as $\mathcal{N}(v)$. Then, for every $l$-th layer with the weight parameter matrix $\mathbf{w}^{(l)}$ and activation function $\sigma(\cdot)$, we can represent its output feature of each node $v$ when connection sampling is applied as the equation \eqref{eq:GDC_sampling} where $\mathbf{w}_{v,u}^{(l)} := \text{diag}(\mathbf{z}_{v,u}^{(l)})\mathbf{w}^{(l)}$ and $\mathbf{h}_u^{(l)}$ denote output feature of node $u$ at $l$-th layer, resulting in learning different random weight parameter $\mathbf{w}_{v,u}^{(l)}$ for each edge $(v, u) \in \mathcal{E}$, i.e., 
\vspace{-0.1cm}
\begin{equation}
    \mathbf{h}_v^{(l+1)} = \sigma \left(\frac{1}{c_v} (\sum_{u \in {\mathcal{N}}(v) \cup \{v\}} \mathbf{z}_{v,u}^{(l)} \odot \mathbf{h}_u^{(l)})\mathbf{w}^{(l)}\right)\\
    = \sigma \left(\frac{1}{c_v} \sum_{u \in {\mathcal{N}}(v) \cup \{v\}} \mathbf{h}_u^{(l)}\mathbf{w}_{v,u}^{(l)}\right).
    \label{eq:GDC_sampling}
\end{equation}
In Bayesian GNNs, we estimate the posterior distribution of GNN weight parameters considering them as random parameters. Since GDC incorporates random parameters including weight parameters $\mathbf{W}^{(l)} = \{\mathbf{w}_e^{(l)}\}_{e=1}^{|\mathcal{E}|}$, random masks $\mathbf{Z}^{(l)} = \{\mathbf{z}_e^{(l)}\}_{e=1}^{|\mathcal{E}|}$, and the corresponding dropout rate $\pi_l$, we infer posteriors on these parameters, denoted as $\boldsymbol{\theta} = \{\mathbf{W}^{(l)}, \mathbf{Z}^{(l)}, \pi_l\}_{l=1}^L$. Note that the drop rates are also learnable to enhance the flexibility of the model \cite{boluki2020learnable}. Then, the free energy criterion for training Bayesian GNNs is defined as 
\begin{equation}
    F_\beta(q) = \mathbb{E}_{q(\boldsymbol{\theta})} \left[\sum_{(x, y) \in \mathcal{D}}-\log p(y|x, \boldsymbol{\theta}) \right] \\
     + \beta \cdot \sum_{l=1}^L \text{KL}\left[q(\boldsymbol{\theta})||p(\boldsymbol{\theta})\right]
\label{GDC VI loss}
\end{equation}
where the \textit{temperature} parameter $\beta$ controls the spread of posterior distribution, as illustrated in Fig.~\ref{fig:framework}. The posterior with $\beta=1$ corresponds to standard Bayesian posterior, and the $\beta \gg 1$ and $\beta \ll 1$ depend more on the prior assumption and information from data, respectively. Consequently, the \textit{tempered posterior} can be obtained by minimizing the equation \eqref{GDC VI loss} as 
\begin{equation}
    q_\beta(\boldsymbol{\theta}|\mathcal{D}) \propto p(\boldsymbol{\theta})\prod_{(x, y)\in \mathcal{D}}p(y|x, \boldsymbol{\theta})^{{1}/{\beta}}.
    \label{tempered posterior}
\end{equation}
Though estimating the posterior distribution in the equation \eqref{tempered posterior} requires intractable computation, \textit{variational inference (VI)} \cite{gal2016dropout} suggests a good approximation for it, where the variational distribution from a defined family of distributions (e.g., Gaussian distributions) approximates true posterior distribution. Details on prior assumptions and VI algorithm for GDC are stated in Appendix \ref{app:BBGDC}. Suppose we have the tempered posterior $q_\beta(\boldsymbol{\theta}|\mathcal{D})$ or its VI approximate, then the predictive distribution on a test sample $x$ can be obtained by averaging over all likely models as
\begin{equation}
     p_\beta(y|x, q) = \int p(y|x, \boldsymbol{\theta}){q}_\beta(\boldsymbol{\theta}|\mathcal{D}) d\boldsymbol{\theta}  \approx \frac{1}{T} \sum_{t=1}^T p(y|x, \hat{\boldsymbol{\theta}}_t) 
     \label{eq:MCsampling}
\end{equation}
where the distribution can be approximated by the average over $T$ sets of i.i.d. samples $\hat{\boldsymbol{\theta}}_t$ drawn from $q_\beta(\boldsymbol{\theta}|\mathcal{D})$. Since we use negative log-loss as non-conformity score for CP, i.e., $-\log{p_\beta(y|x, q)}$, the prediction sets, especially the size of sets is directly affected by the temperature. Accordingly, we discuss the adjustment of temperature of Bayesian GNNs for CP in the following. 

\begin{figure}[!t]
     \centering
    \begin{subfigure}[t]{0.93\textwidth}
        \raisebox{-\height}{\includegraphics[width=0.31\textwidth]{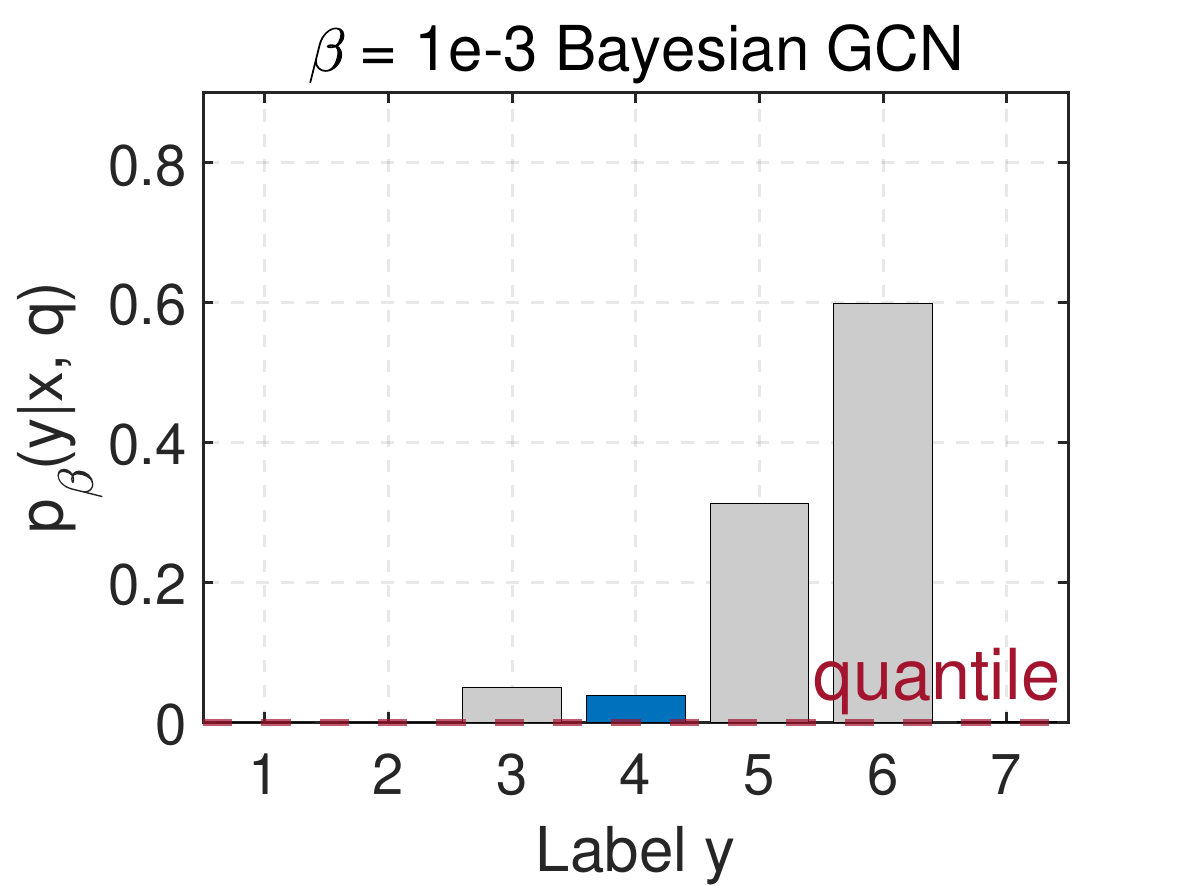}}
        \raisebox{-\height}{\includegraphics[width=0.31\textwidth]{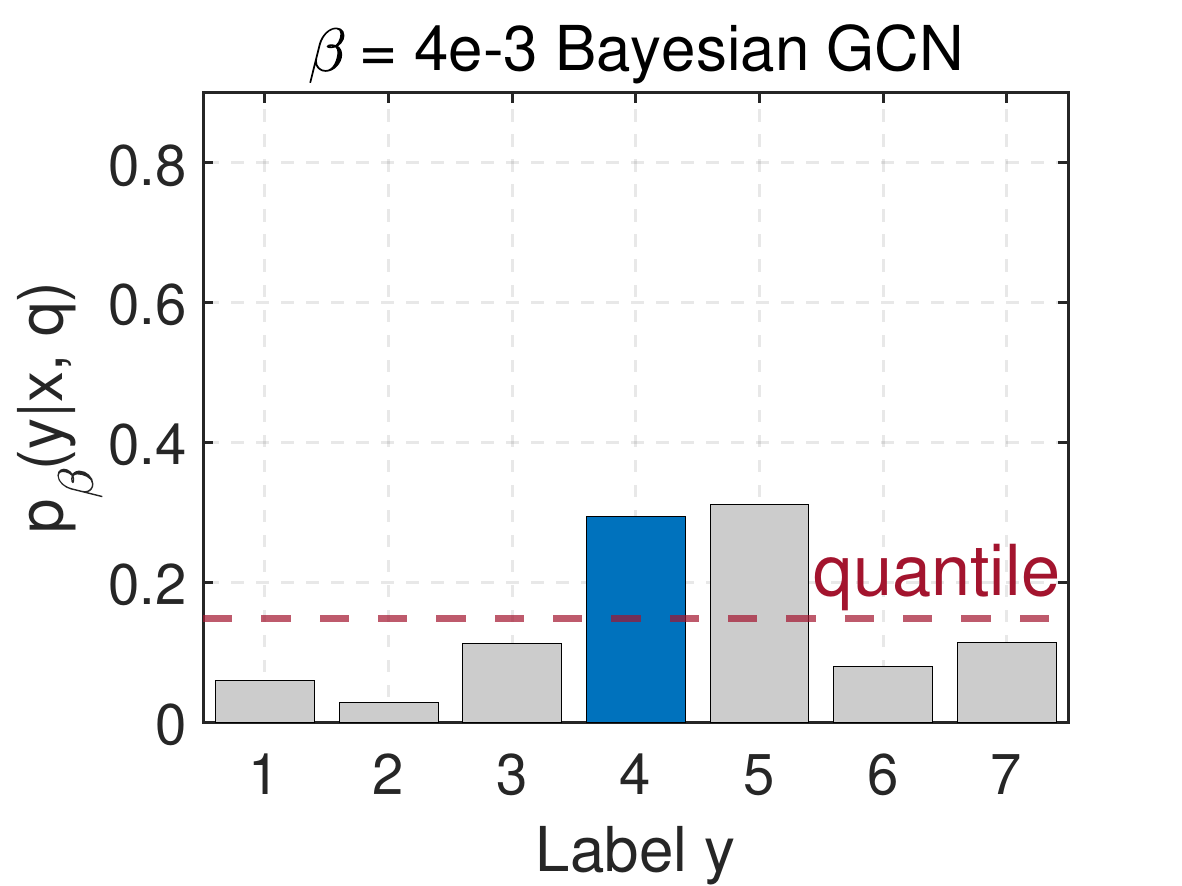}}
        \raisebox{-\height}{\includegraphics[width=0.31\textwidth]{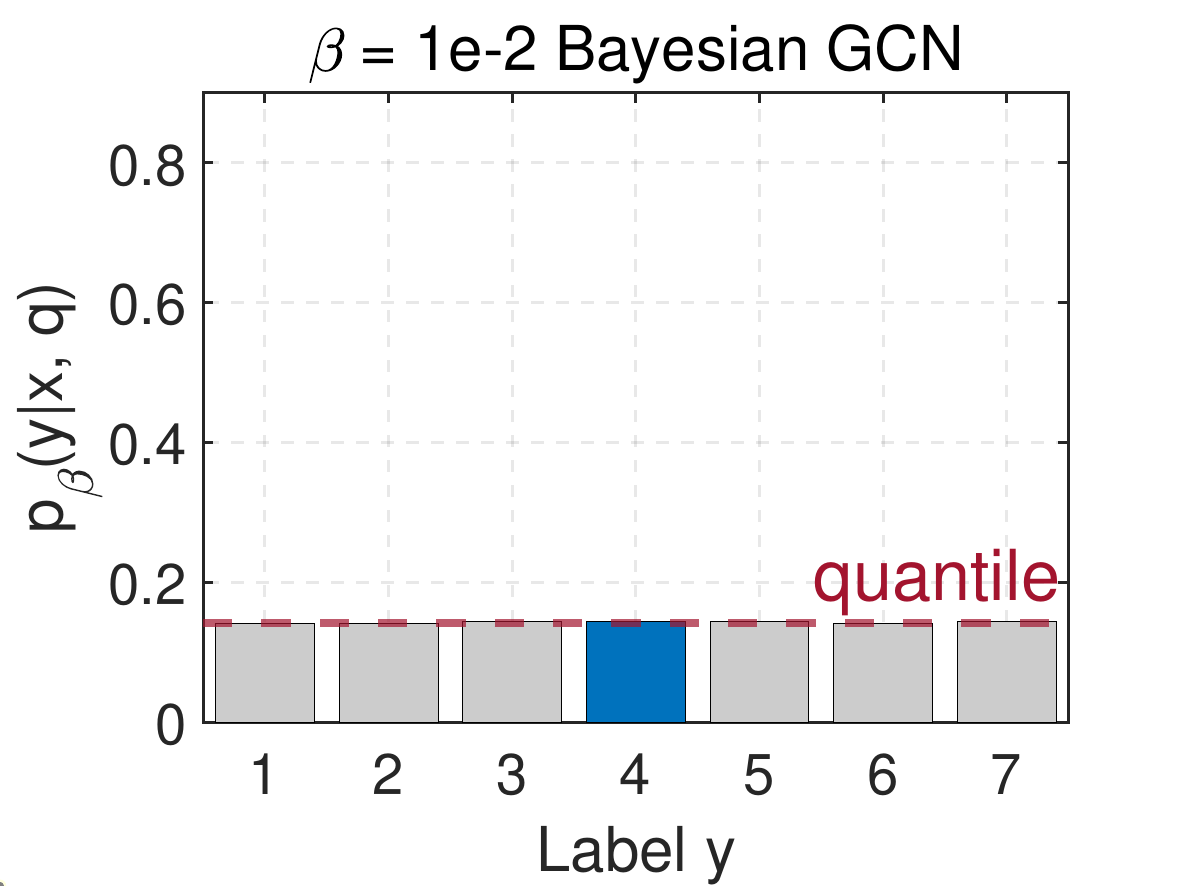}}
          \caption{Label distribution on Cora dataset. }
    \end{subfigure}
        \begin{subfigure}[t]{0.93\textwidth}
        \raisebox{-\height}{\includegraphics[width=0.31\textwidth]{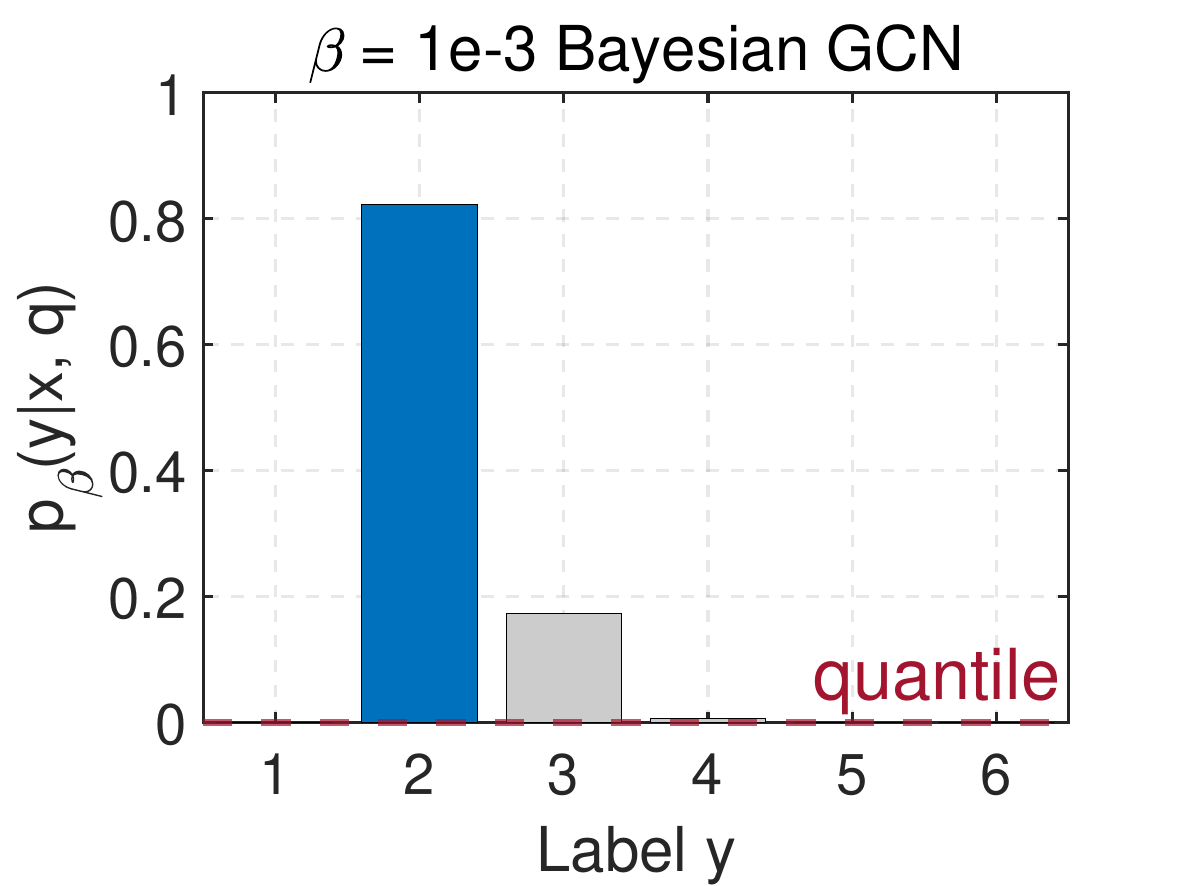}}
        \raisebox{-\height}{\includegraphics[width=0.31\textwidth]{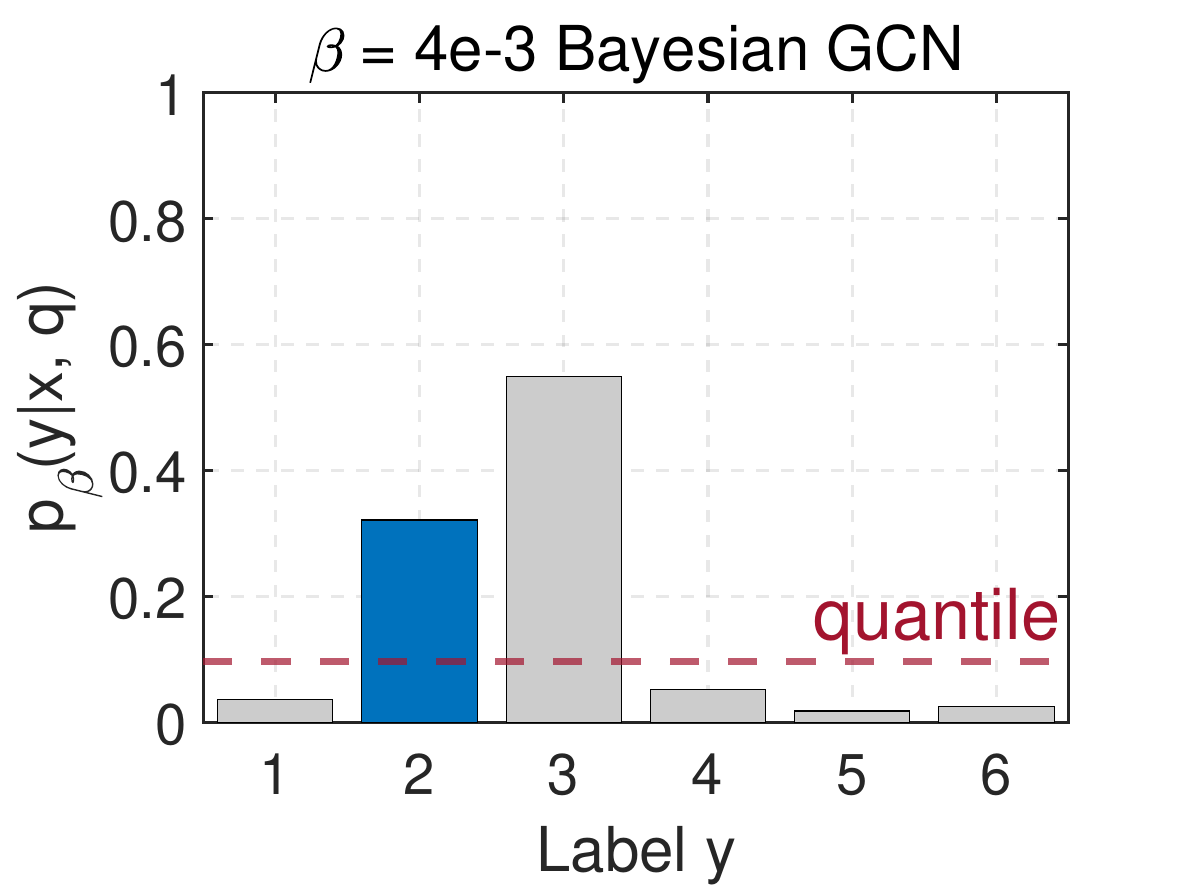}}
        \raisebox{-\height}{\includegraphics[width=0.31\textwidth]{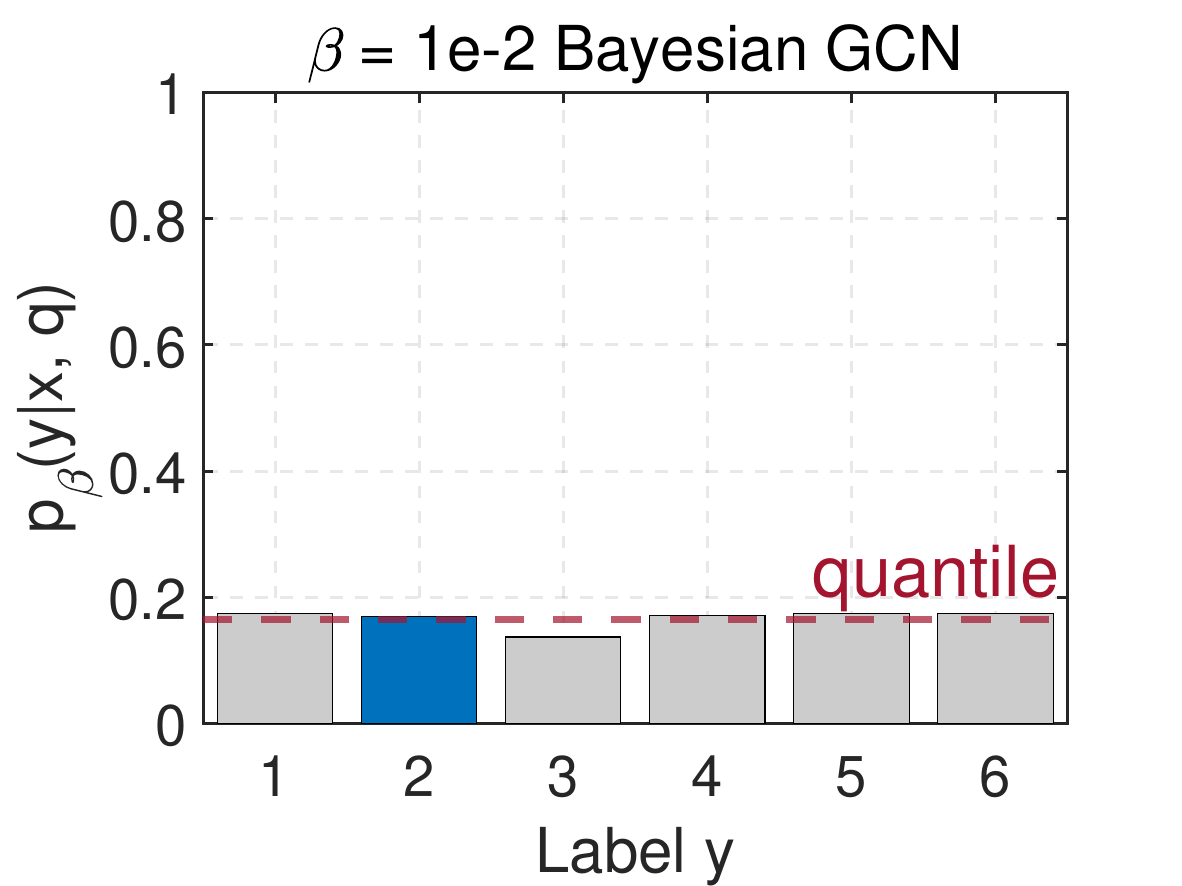}}
      \caption{Label distribution on Citeseer dataset. }
    \end{subfigure}
    \caption{\textbf{Example of label distribution depending on the temperature of Bayesian GCNs.}}
    \label{fig:label dist}
\end{figure}

\subsection{Temperature Control for Conformal Prediction}

We now present how to exploit the benefits of Bayesian GNNs with tempered posterior in split conformal prediction (SCP). As illustrated in Fig.~\ref{fig:framework}, we first train Bayesian GNNs using loss function in the equation \eqref{GDC VI loss} with a pre-determined temperature $\beta$ using training set. Afterward, for a given calibration set $\{(x_i, y_i)\}_{i=1}^n$, we compute the non-conformity score of calibration data as negative log-loss based on the predictive distribution obtained from the tempered posterior as the equation \eqref{eq:MCsampling}. Subsequently, for a given pre-defined coverage level of $1-\alpha$, the prediction set for a test input $x$, generated using SCP on top of Bayesian GNNs with tempered posterior, is defined as follows:  
\begin{equation}
    \mathcal{C}_{n, \alpha, \beta}(x) =\left\{y\in\mathcal{Y}: -\log{p_\beta(y|x, q)} \le Q_{\alpha}(\{-\log{p_\beta(y_i|x_i, q)}\}_{i=1}^n)\right\}.
    \label{eq:CPset_beta} 
\end{equation}
This prediction set satisfies the desired coverage condition, as stated in \autoref{thm2: gnn_cp_cov}. Additionally, as depicted in Fig.~\ref{fig:framework}, we suggest that employing the tempered posterior in Bayesian GNNs, where the temperature controls its spread as means to overcome model misspecification, can generate an efficient prediction set within CP. While the credible region introduced in \cite{syring2019calibrating} provides a coverage guarantee in an asymptotic sense, SCP provides a rigorous guarantee without relying on specific distribution assumptions, as outlined in \autoref{thm1: cov}. Given that conformal predictors alone may not necessarily create the smallest prediction sets, adjusting the temperature provides greater freedom to reduce the size of the prediction sets. Please refer to Appendix \ref{app:pf} for the proofs of the theorems. 
\begin{theorem} 
    Given that calibration set $\{(x_i,y_i)\}_{i=1}^n$ and a test data point $(x,y)$ are exchangeable random variables, for any coverage level $1-\alpha$, and for any pre-determined temperature parameter $\beta \in [0, \infty)$, a prediction set $\mathcal{C}_{n, \alpha, \beta}(x)$ defined in the equation \eqref{eq:CPset_beta} satisfies the equation \eqref{eq1} with additional marginalization over posterior sampling in the equation \eqref{eq:MCsampling} when assuming a finite number of sampling $T$. 
    \label{thm2: gnn_cp_cov}
\end{theorem}

To suggest a brief insight into how the temperature affects the inefficiency, we provide an example of label distribution from Bayeisan GNNs at different temperatures, extracted from real data samples of Cora and Citeseer. In Fig.~\ref{fig:label dist}, note that there are some cases where the true label is included in prediction set, but not selected as a point prediction. While prediction sets from all models include the true label, the size largely depends on the temperature of Bayesian GNNs. As depicted in Fig.~\ref{fig:framework}, when the temperature parameter $\beta$ is set too low, the label distribution tends to reflect an over-confident behavior compared to other models. This leads to extremely low quantiles and prediction set containing a large number of candidate labels. Conversely, if the temperature is set too high, it results in labels having nearly identical confidence levels, leading to prediction sets with less informative labels. However, when an appropriate temperature is selected between these extremes, Bayesian GNNs generate efficient prediction sets that contain fewer number of informative labels, resulting in better efficiency.

\vspace{-0.1cm}
\section{Experimental Results}
In this section, we study (i) the conformal prediction performance of tempered posterior in GCN, in terms of empirical coverage and inefficiency, and analyze (ii) the contributing factors of inefficiency, considering the connection between inefficiency and model calibration.

\begin{table*}[!b]
  \centering
  \begin{tabular}{ccccccccc}
    \toprule
    Dataset & \#Nodes & \#Edges & \#Graphs &  \#Feat & \#Classes & \#Train & \#Cal & \#Test \\
    \midrule
    Cora & 2708 & 5429 & - & 1433 & 7 & 140 & 500 & 1000\\
    Citeseer & 3327 & 4732 & - & 3703 & 6 & 120 & 500 & 1000\\
    CIFAR10 & - & - & 50000 & 5 & 10 & 10000 & 5000 & 10000\\
    ENZYMES  & - & - & 600 & 21 & 6 & 500 & 50 & 50 \\
    \bottomrule
  \end{tabular}
  \caption{\label{tab1:data} Dataset statistics for Cora, Citeseer, CIFAR10, ENZYMES}
\end{table*}

\paragraph{Experimental settings} We consider two major tasks on graph-structured data, which are graph classification and node classification tasks. We focus on transductive setting in node classification task that ensures the validity \cite{huang2023uncertainty}. We have selected public graph datasets with enough number of classes that can effectively reflect the change of set size depending on the temperature. We use CIFAR10-superpixel \cite{knyazev2019understanding, dwivedi2020benchmarking} and ENZYMES \cite{morris2020tudataset} for graph classification task, and Cora and Citeseer for node classification task \cite{sen2008collective}. Overall dataset statistics used for model training and CP are stated in Table \ref{tab1:data}. Our models are based on GCN \cite{kipf2016semi} and Bayesian GCNs are trained with Graph DropConnect (GDC) assuming the same priors for random parameters as \cite{hasanzadeh2020bayesian}. We have selected temperature parameters for each dataset that make reasonable test accuracy. To ensure the exchangeability of split conformal prediction, we randomly split calibration and test set as illustrated in Table \ref{tab1:data}, and CP results are averaged over 50 runs for CIFAR10, and 100 runs for other datasets. Further details on experimental settings and exchangeability conditions for GNNs can be seen in Appendix \ref{app:exch}, \ref{app:exp}.

\paragraph{Evaluation metrics} We use two metrics of conformal prediction, empirical coverage and empirical inefficiency. A conformal predictor is considered to be valid when the empirical coverage over test set $\mathcal{D}_\text{test}$, denoted as $\widehat{\textit{cov}}$, satisfies the pre-defined coverage level (we use 0.9). Note that achieving higher coverage is not necessarily advantageous if it leads to an increase in set size, quantified as higher inefficiency. Another metric is empirical inefficiency which measures the average size of prediction sets over test set, denoted as $\widehat{\textit{ineff}}$. Lower inefficiency indicates that a conformal predictor is more efficient since it outputs smaller number of candidate labels while keeping the same pre-defined coverage level. 
\begin{equation}
    \widehat{\textit{cov}}=\frac{1}{\left\vert\mathcal{D}_\text{test}\right\vert}\sum_{(x, y) \in \mathcal{D}_\text{test}} \mathbbm{1}\{y \in C_{n, \alpha} (x) \}, \,\, 
    \widehat{\textit{ineff}}=\frac{1}{\left\vert\mathcal{D}_\text{test}\right\vert}\sum_{(x, y) \in \mathcal{D}_\text{test}} \left\vert C_{n, \alpha} (x)\right\vert
\end{equation}

\paragraph{Calibration measures} To assess the calibration of GCNs, we employ commonly used calibration measures, along with CP measures. For each data sample $(x, y)$, suppose we have a point prediction from a model $\theta$ as $\hat{y} = \argmax_c p(c|x, \theta)$, then accuracy and confidence for every sample are defined as $\mathbbm{1}\{\hat{y}=y\}$ and $p(c=\hat{y}|x, \theta)$, respectively. We use reliability diagram, which depicts the average sample accuracy against confidence levels. If a model is perfectly calibrated, where the estimated probability of true label is exactly the same as the confidence for all levels, the reliability diagram takes the form of an identity function. In our experiments, we divide samples into $M=20$ confidence bins, each denoted as $B_m$, whose samples have confidence within the range $(\frac{m-1}{M}, \frac{m}{M}]$. The average accuracy and confidence over samples in $B_m$ are denoted as $\text{acc}(B_m)$ and $\text{conf}(B_m)$, respectively. 
We additionally use two measures obtained from reliability diagram, expected calibration error (ECE) and maximum calibration error (MCE) \cite{guo2017calibration}, defined as 
\begin{equation}
    \text{ECE} = \sum_{m=1}^M \frac{\left\vert{B_m}\right\vert}{n} \left\vert \text{acc}(B_m)-\text{conf}(B_m)\right\vert, \; \text{MCE} = \max_{m \in \{1,\cdots, M\}}  \left\vert \text{acc}(B_m)-\text{conf}(B_m)\right\vert,
\end{equation}
where $n$ is the number of total samples. ECE is a weighted average of differences between confidence and accuracy across all bins, while MCE represents the largest of these differences among all bins. 

\subsection{Efficiency of Tempered Posteriors}

\begin{figure}
     \centering
    \begin{subfigure}[t]{0.245\textwidth}
        \raisebox{-\height}{\includegraphics[width=\textwidth]{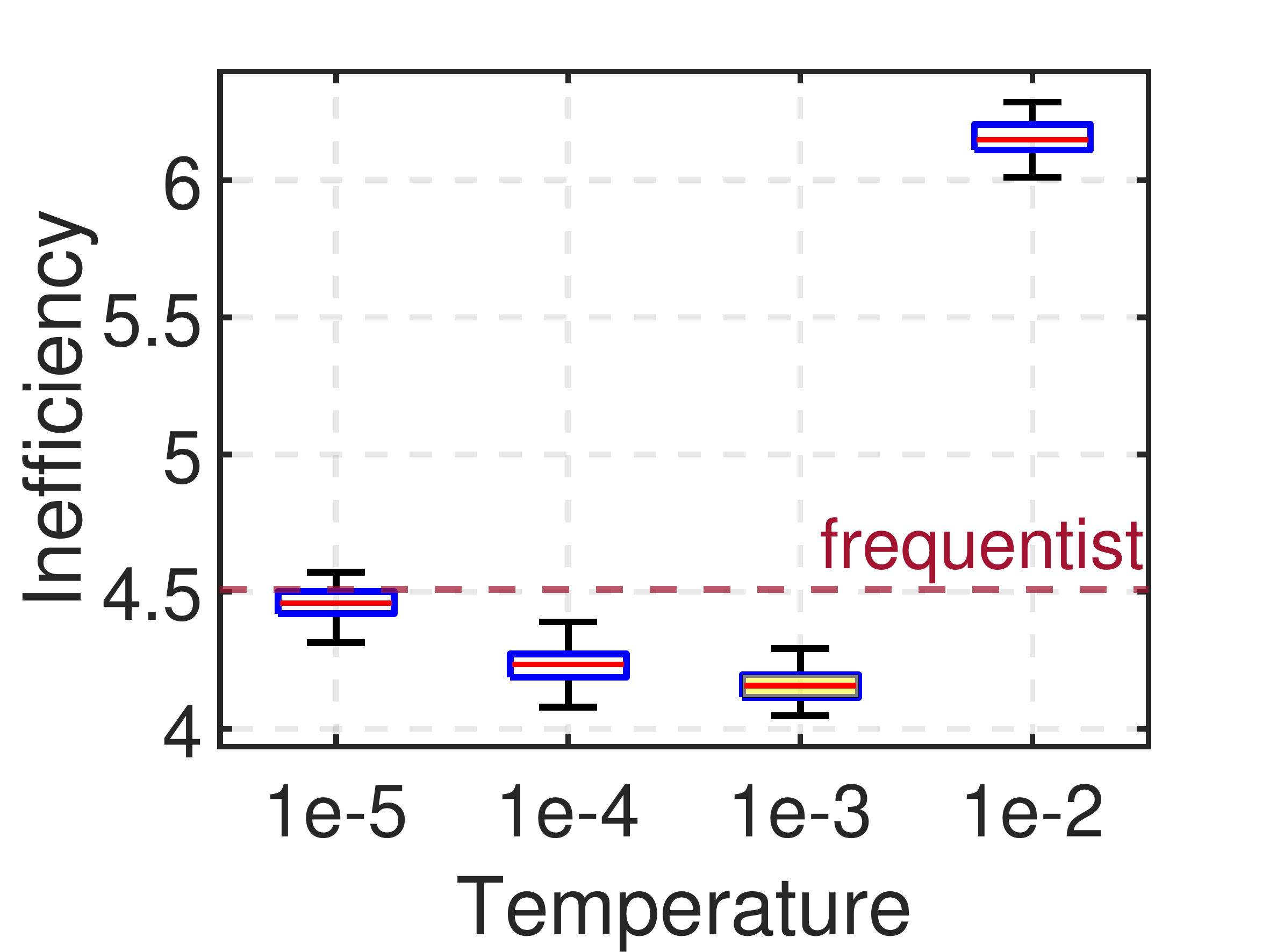}}
    \end{subfigure}
    \hfill
    \begin{subfigure}[t]{0.245\textwidth}
        \raisebox{-\height}{\includegraphics[width=\textwidth]{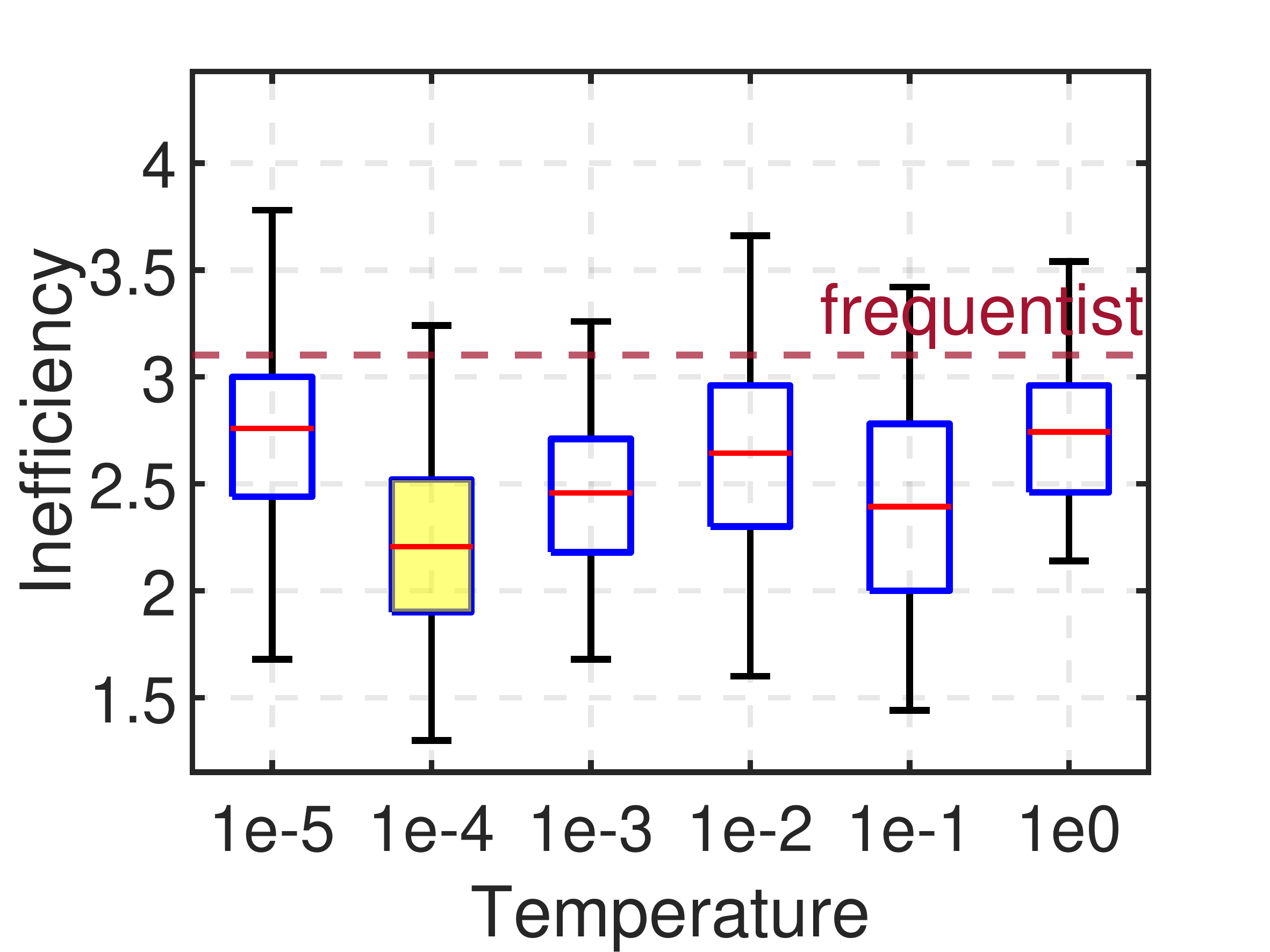}}
    \end{subfigure}
    \hfill
    \begin{subfigure}[t]{0.245\textwidth}
        \raisebox{-\height}{\includegraphics[width=\textwidth]{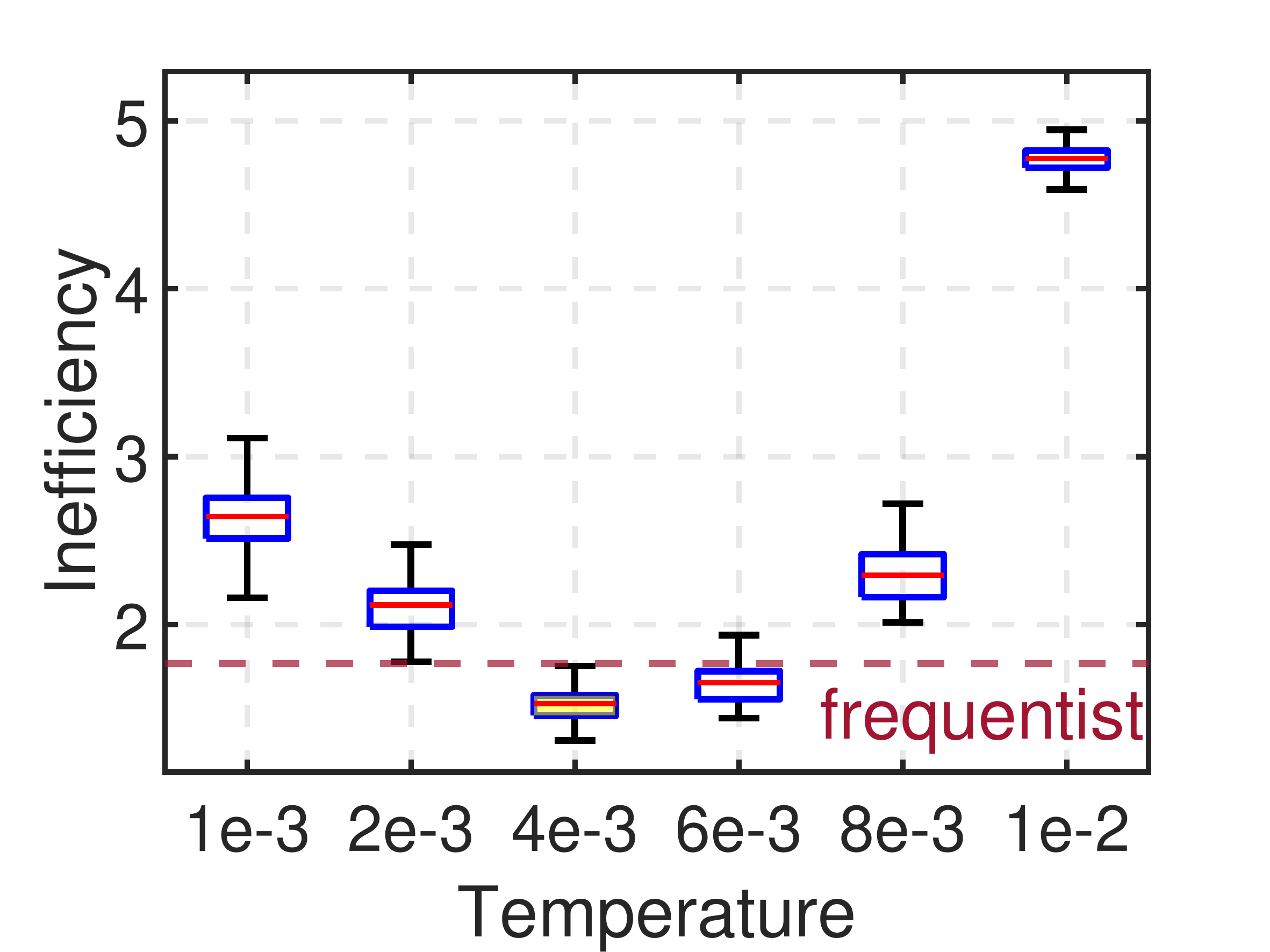}}
    \end{subfigure}
    \hfill
    \begin{subfigure}[t]{0.245\textwidth}
        \raisebox{-\height}{\includegraphics[width=\textwidth]{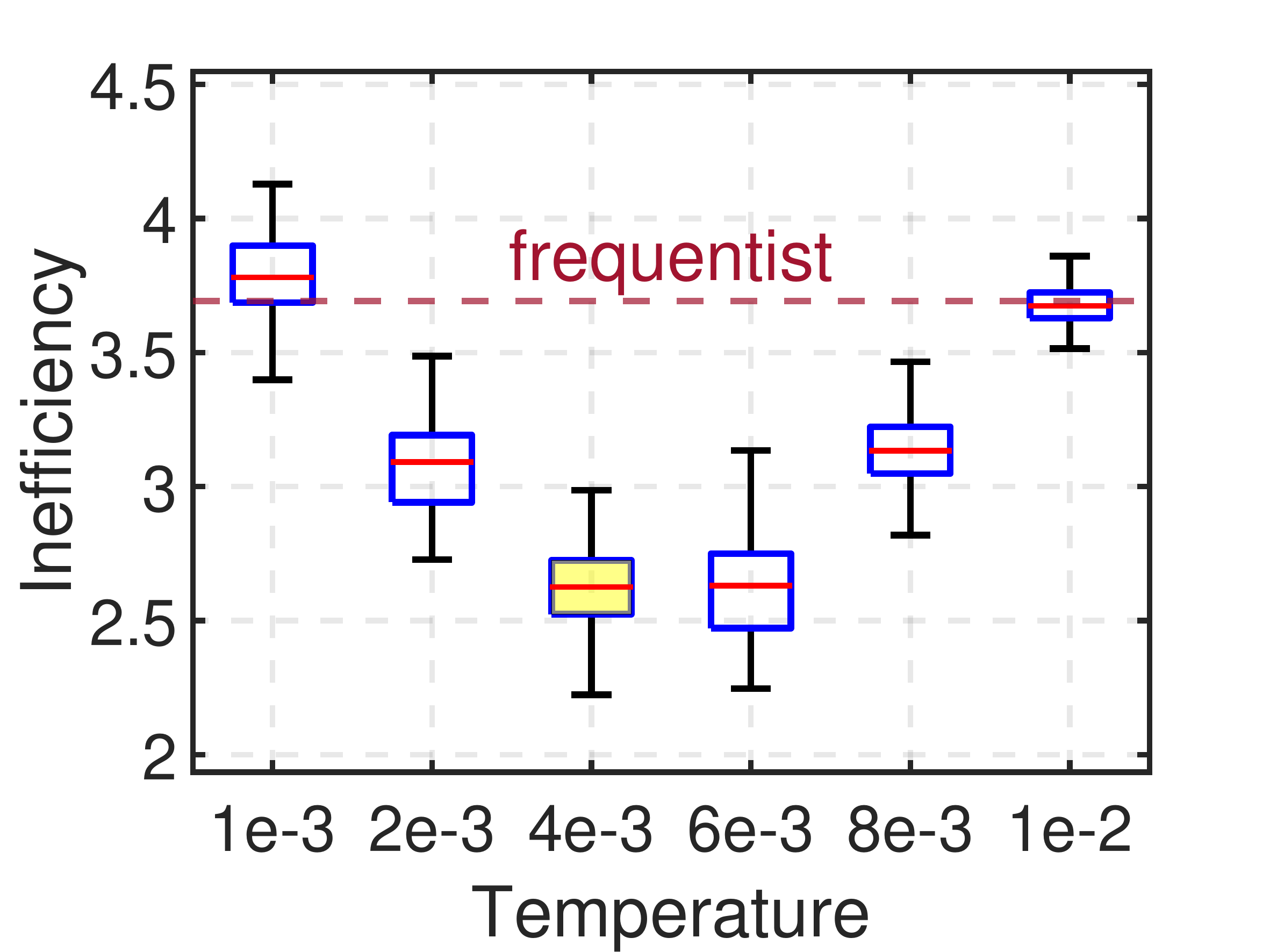}}
    \end{subfigure}
    \begin{subfigure}[t]{0.245\textwidth}
        \raisebox{-\height}{\includegraphics[width=\textwidth]{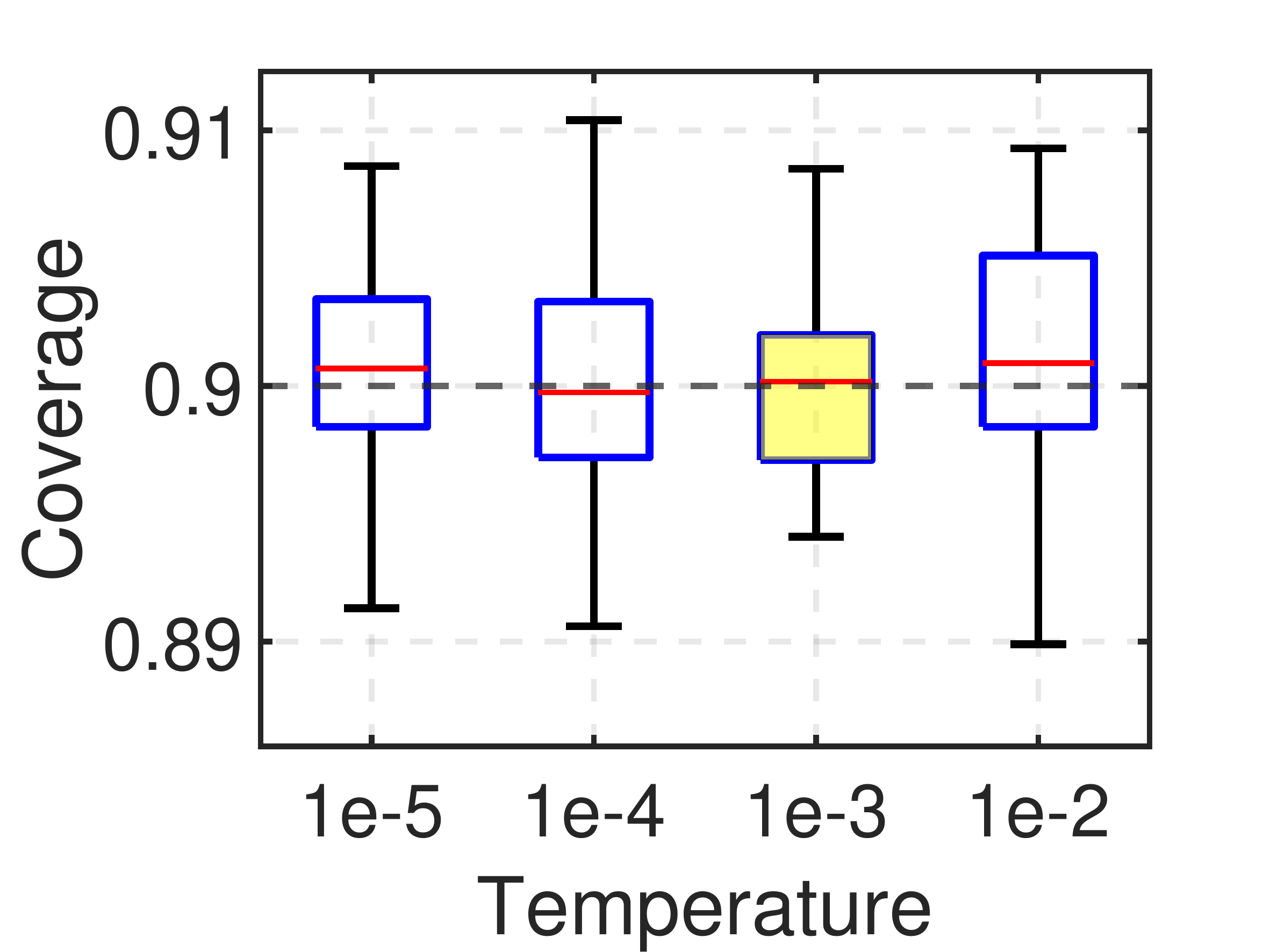}}
        \caption{CIFAR10-superpixel}
    \end{subfigure}
    \hfill
    \begin{subfigure}[t]{0.245\textwidth}
        \raisebox{-\height}{\includegraphics[width=\textwidth]{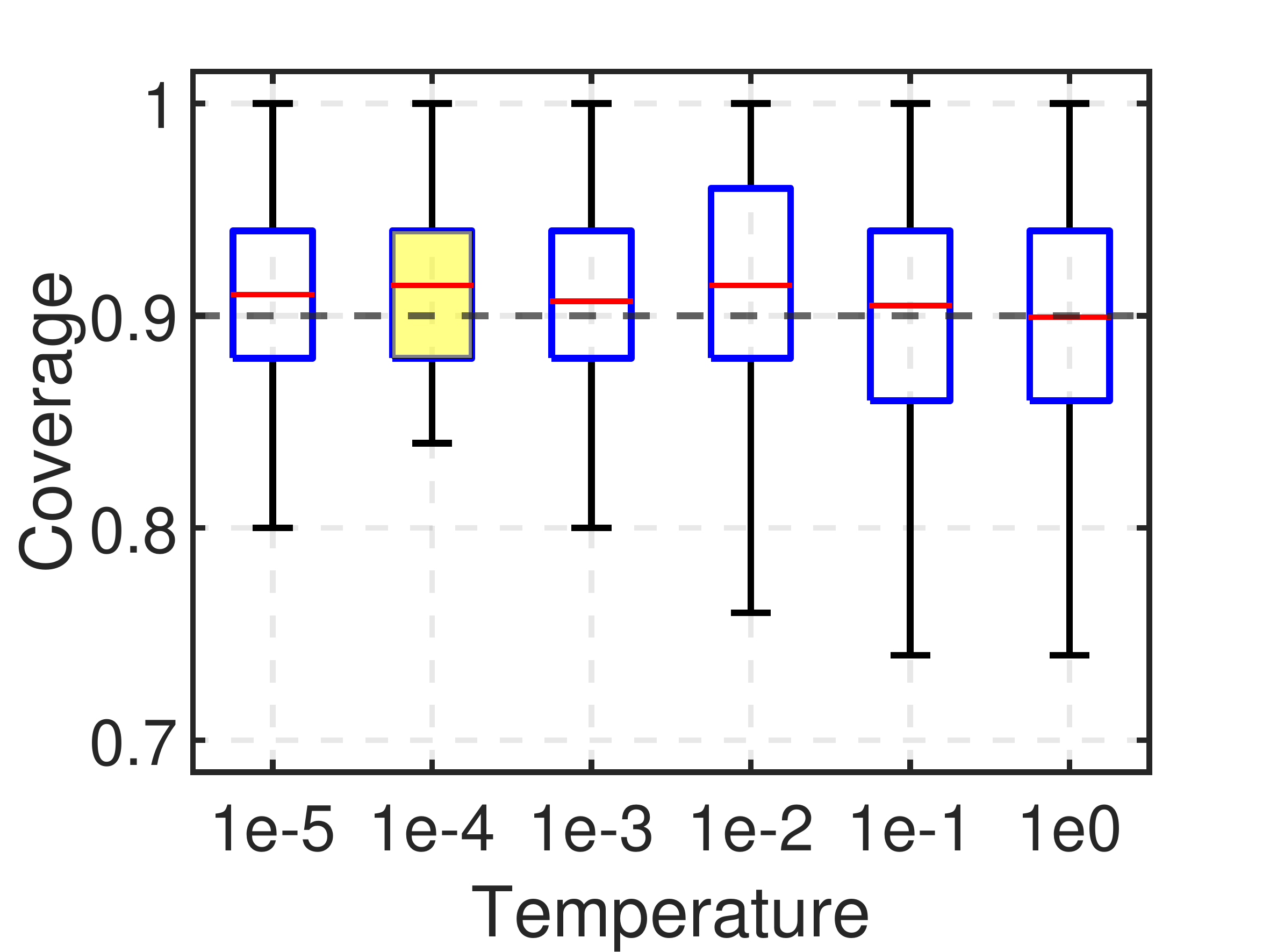}}
        \caption{ENZYMES}
    \end{subfigure}
    \hfill
    \begin{subfigure}[t]{0.245\textwidth}
        \raisebox{-\height}{\includegraphics[width=\textwidth]{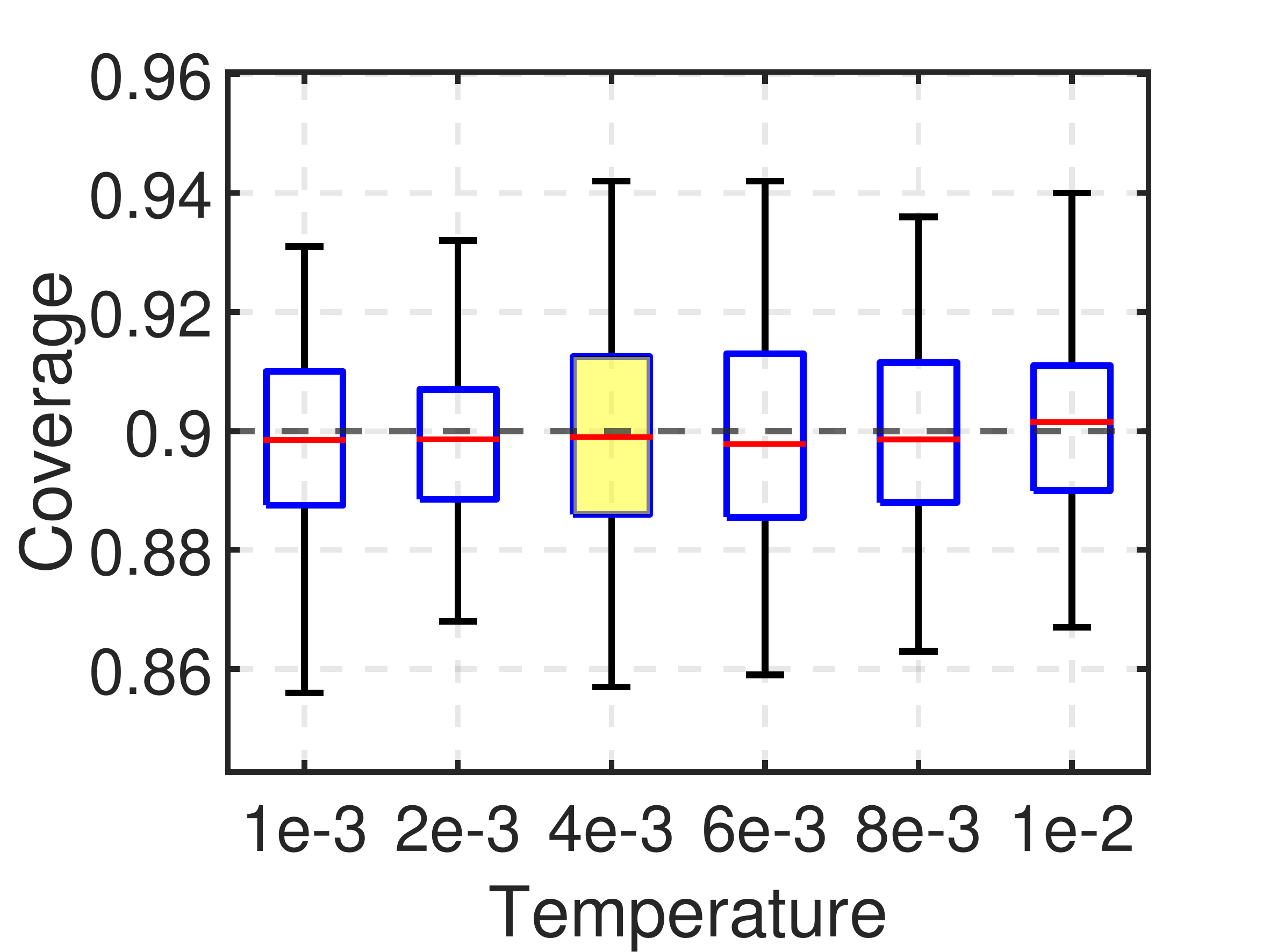}}
        \caption{Cora}
    \end{subfigure}
    \hfill
    \begin{subfigure}[t]{0.245\textwidth}
        \raisebox{-\height}{\includegraphics[width=\textwidth]{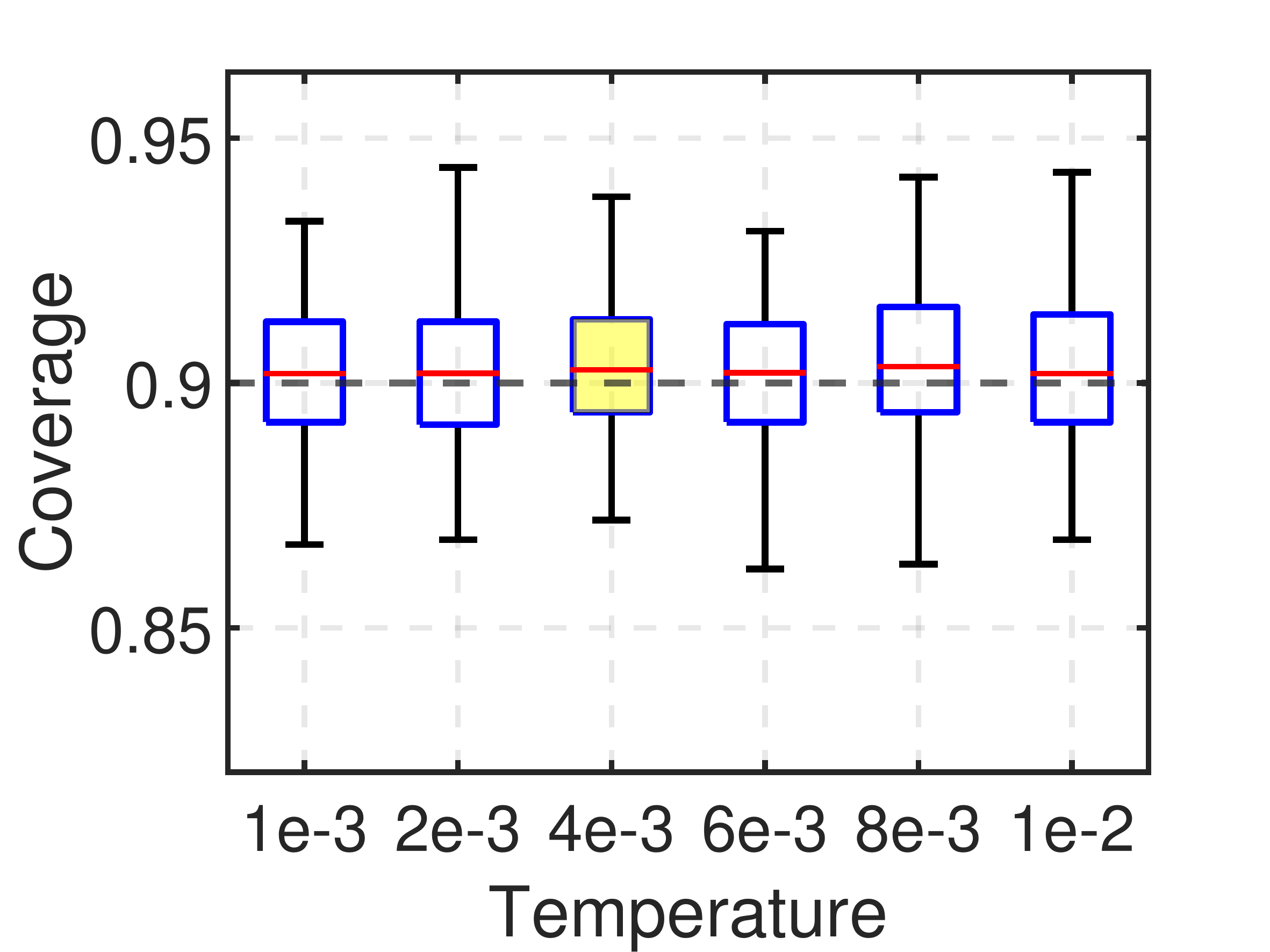}}
        \caption{Citeseer}
    \end{subfigure}
    \caption{\textbf{Empirical coverage and inefficiency of Bayesian GCNs with different temperatures}. Model that produces smallest sized prediction sets on average is denoted as yellow box. The exact average coverage and inefficiency of Bayesian models are represented as red line inside each box.}
    \label{fig:CP perf}
\end{figure}

We evaluated our frequentist and Bayesian GCNs using CP evaluation metrics, empirical coverage and empirical inefficiency in Fig.~\ref{fig:CP perf}. As theoretically guaranteed, Bayesian models of all temperatures satisfy the pre-defined coverage level of 0.9. Since coverage is guaranteed empirically across all models, the empirical inefficiency determines the performance of conformal predictors. A conformal predictor producing a smaller prediction set implies that the set conveys valuable information. In Fig.~\ref{fig:CP perf}, we have verified that Bayesian GCN gives benefits to the lower inefficiency, while guaranteeing the desired coverage. {\it Notably, there exists a model with a temperature that makes better inefficiency, compared to other temperatures}, as illustrated as yellow boxes in Fig.~\ref{fig:CP perf}. One may question that model with high accuracy might create smaller sets, and results in lower inefficiency. However, {it is worth noting that} while the inefficiency gets affected by the model performance, {we have observed that} this is not always the case. In the following sections, we will delve into this matter further by examining the relationship between inefficiency and model calibration.

\subsection{Contributing Factors to Inefficiency}
\label{results:cal}

\begin{figure}[!t]
     \centering
    \begin{subfigure}[t]{1\textwidth}
        \raisebox{-\height}{
        
        \includegraphics[width=0.32\textwidth]{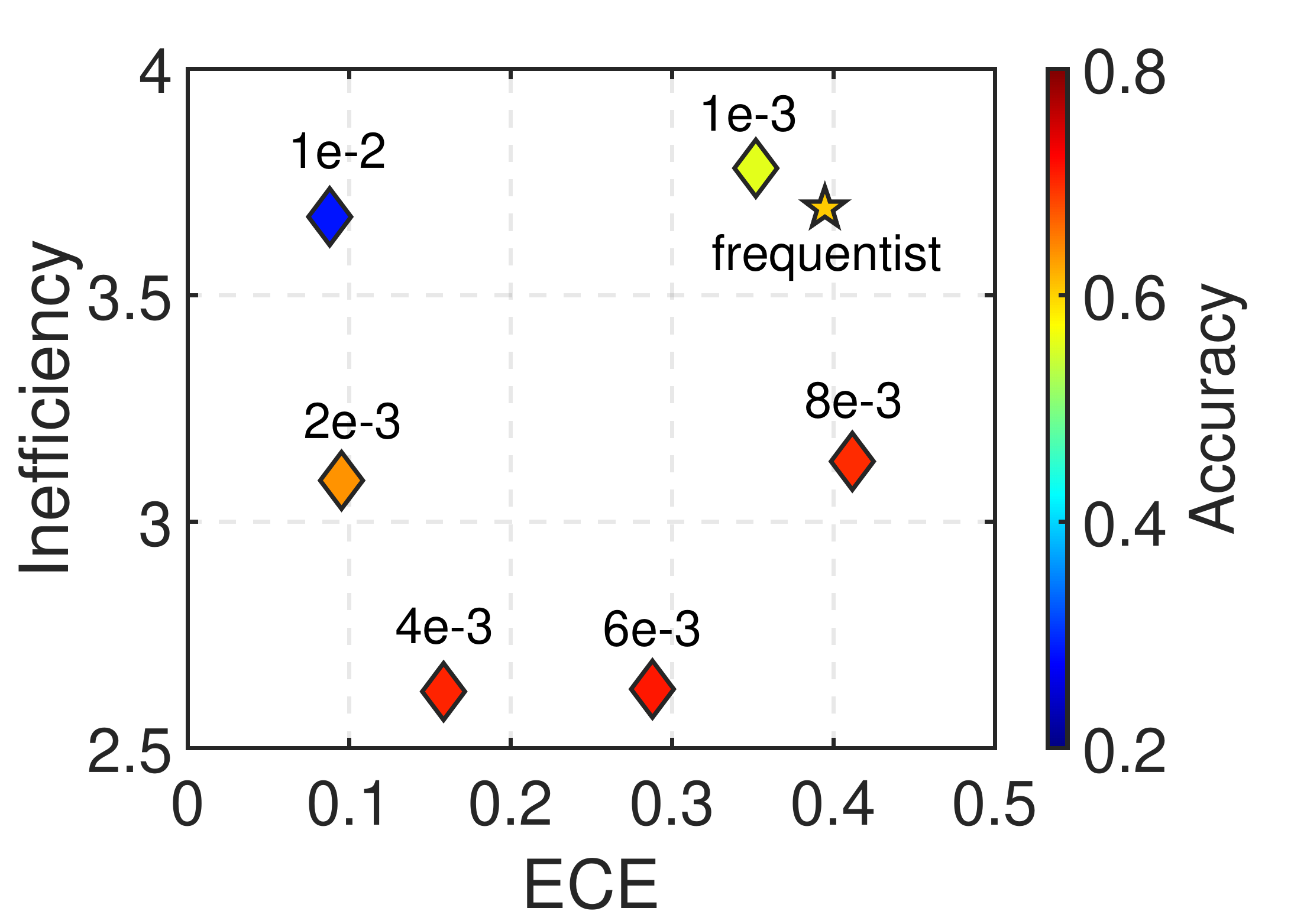}}
        \raisebox{-\height}{\includegraphics[width=0.32\textwidth]{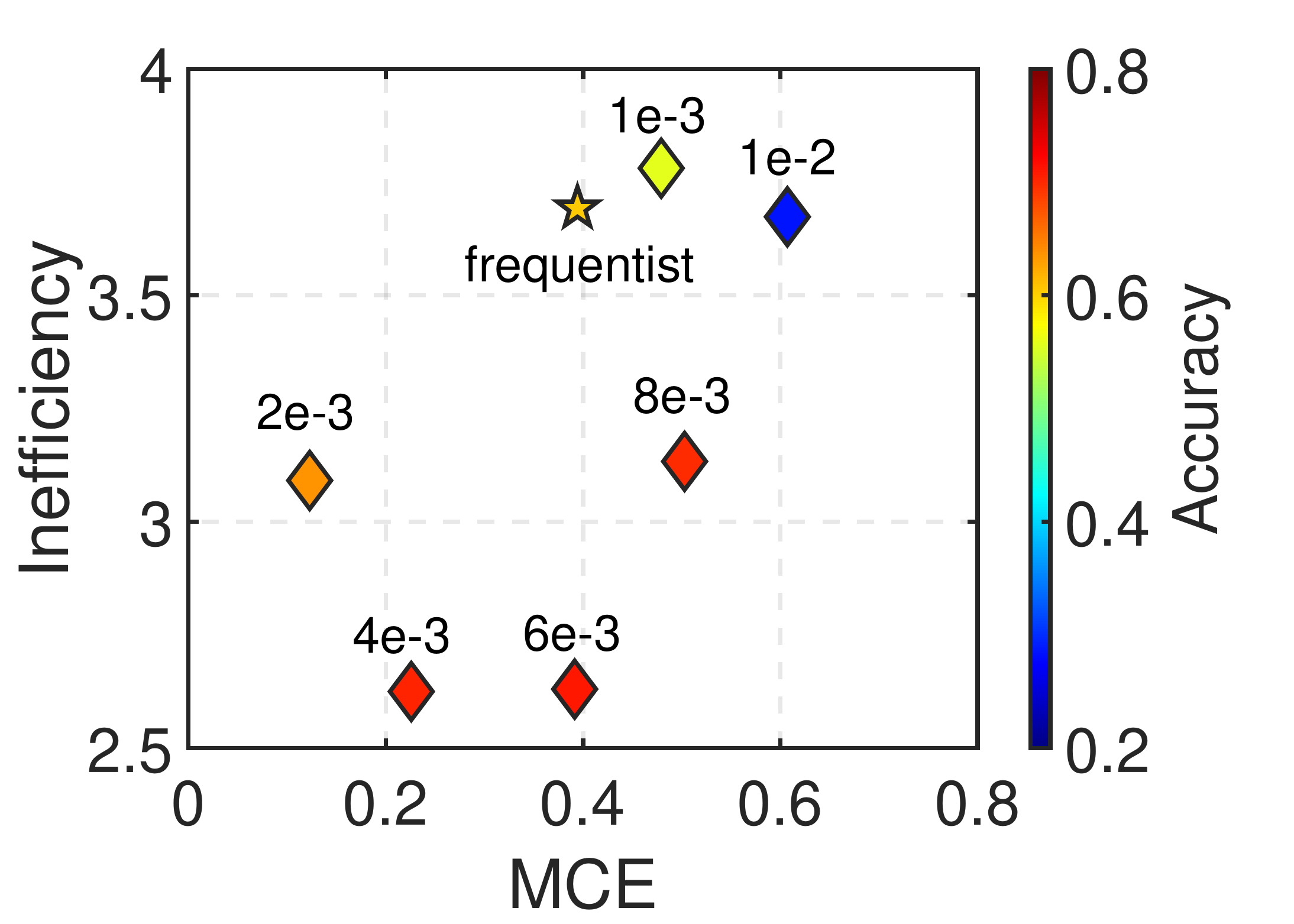}}
        \raisebox{-\height}{\includegraphics[width=0.32\textwidth]{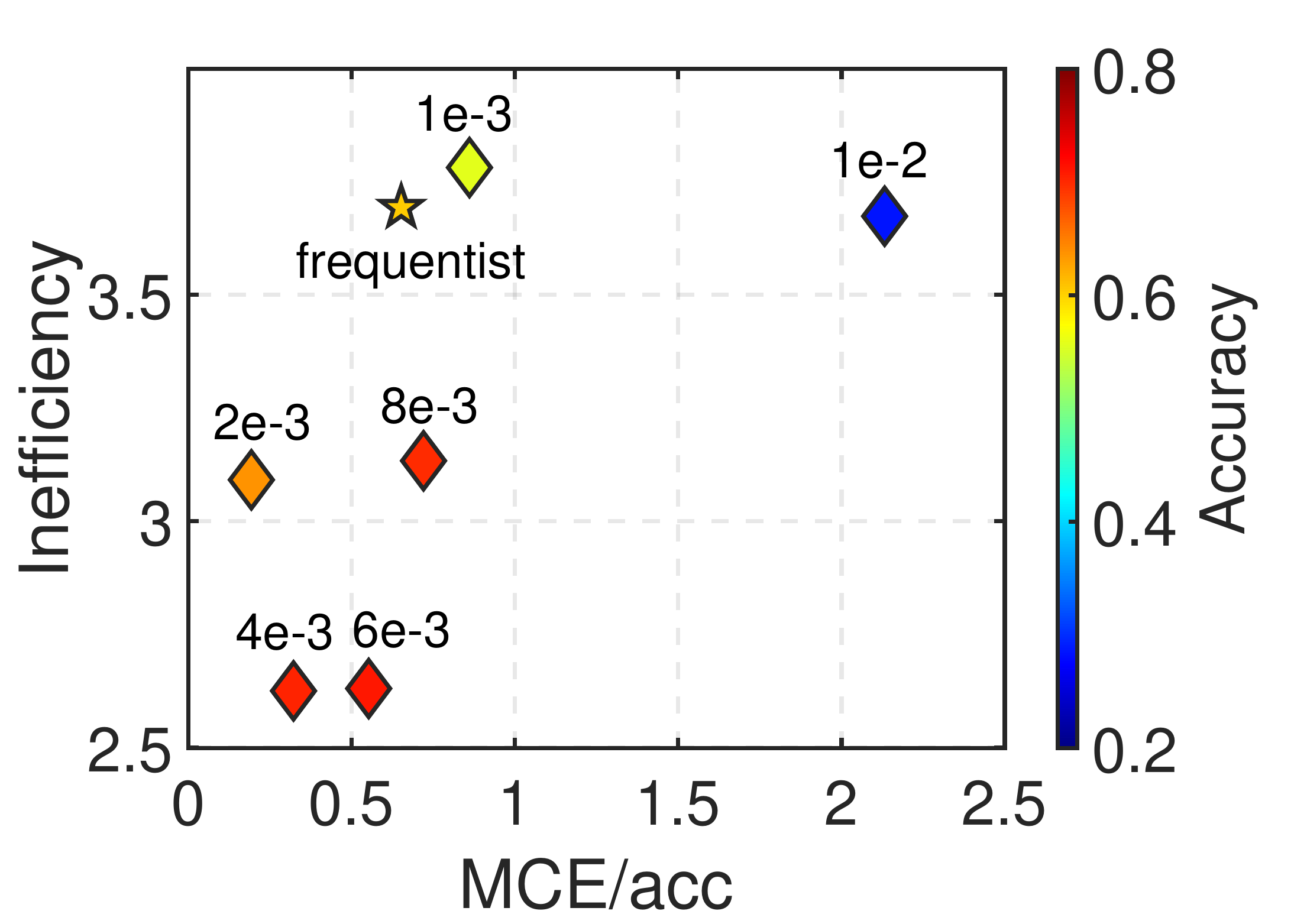}}
        \caption{Relation between inefficiency and calibration measures}
        \vspace{0.3cm}
        \label{subfig:cite-a}
    \end{subfigure}
    \begin{subfigure}[t]{1\textwidth}
        \raisebox{-\height}{\includegraphics[width=0.32\textwidth]{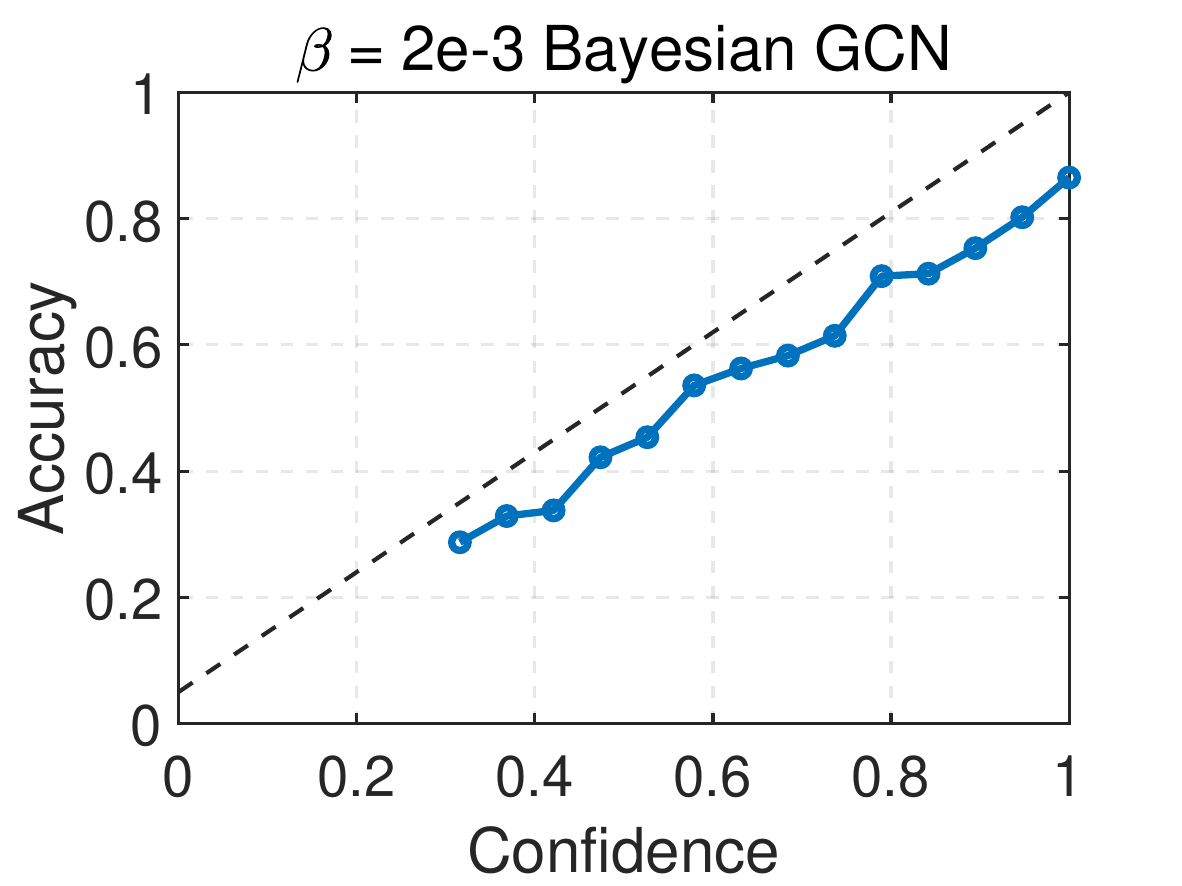}}
        \raisebox{-\height}{\includegraphics[width=0.32\textwidth]{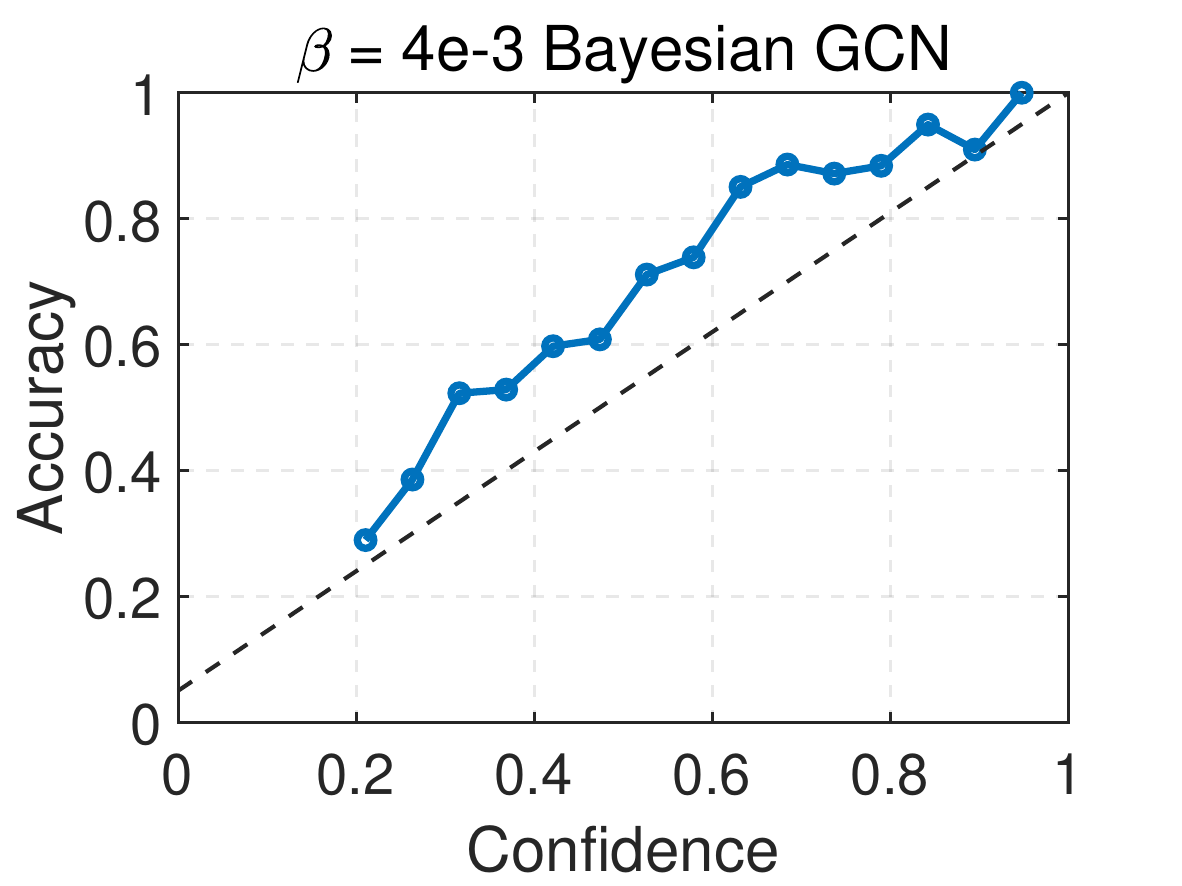}}
        \raisebox{-\height}{\includegraphics[width=0.32\textwidth]{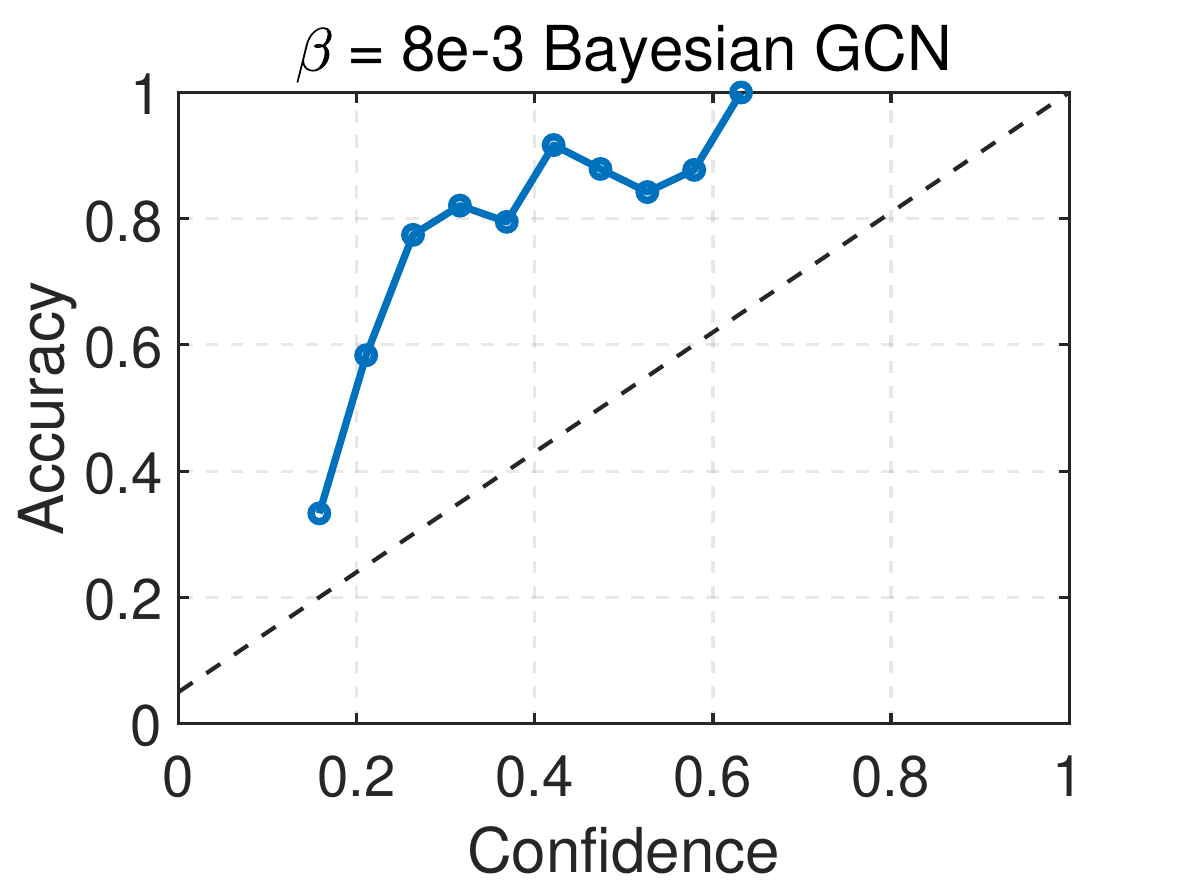}}
        \caption{Reliability diagrams of trained Bayesian GCNs on Citeseer}
        \label{subfig:cite-b}
    \end{subfigure}
  \caption{\textbf{Results of inefficiency and calibration on Citeseer}. (a) Model with lower inefficiency implies that it generates prediction sets with smaller size on average. Bayesian GCNs with $\beta=4 \cdot10^{-3} $ has the lowest inefficiency. (b) The line $y=x$ on reliability diagrams represents a perfectly calibrated model.}
  \label{fig:cite reliability}
\end{figure}

Our experiments have shown that inefficiency is influenced by not only model performance but also model calibration. For example in Fig.~\ref{subfig:cite-a}, the Bayesian model with the temperature $\beta=2\cdot10^{-3}$ has lower inefficiency than the model with $\beta=8\cdot10^{-3}$, even though the test accuracy is lower. Motivated by this observation, we conducted following analysis to answer to the quenstion \textit{“What factors contribute to inefficiency in CP due to the changes in temperature?”}

\begin{figure}[!t]
     \centering
    \begin{subfigure}[t]{1\textwidth}
        \raisebox{-\height}{\includegraphics[width=0.32\textwidth]{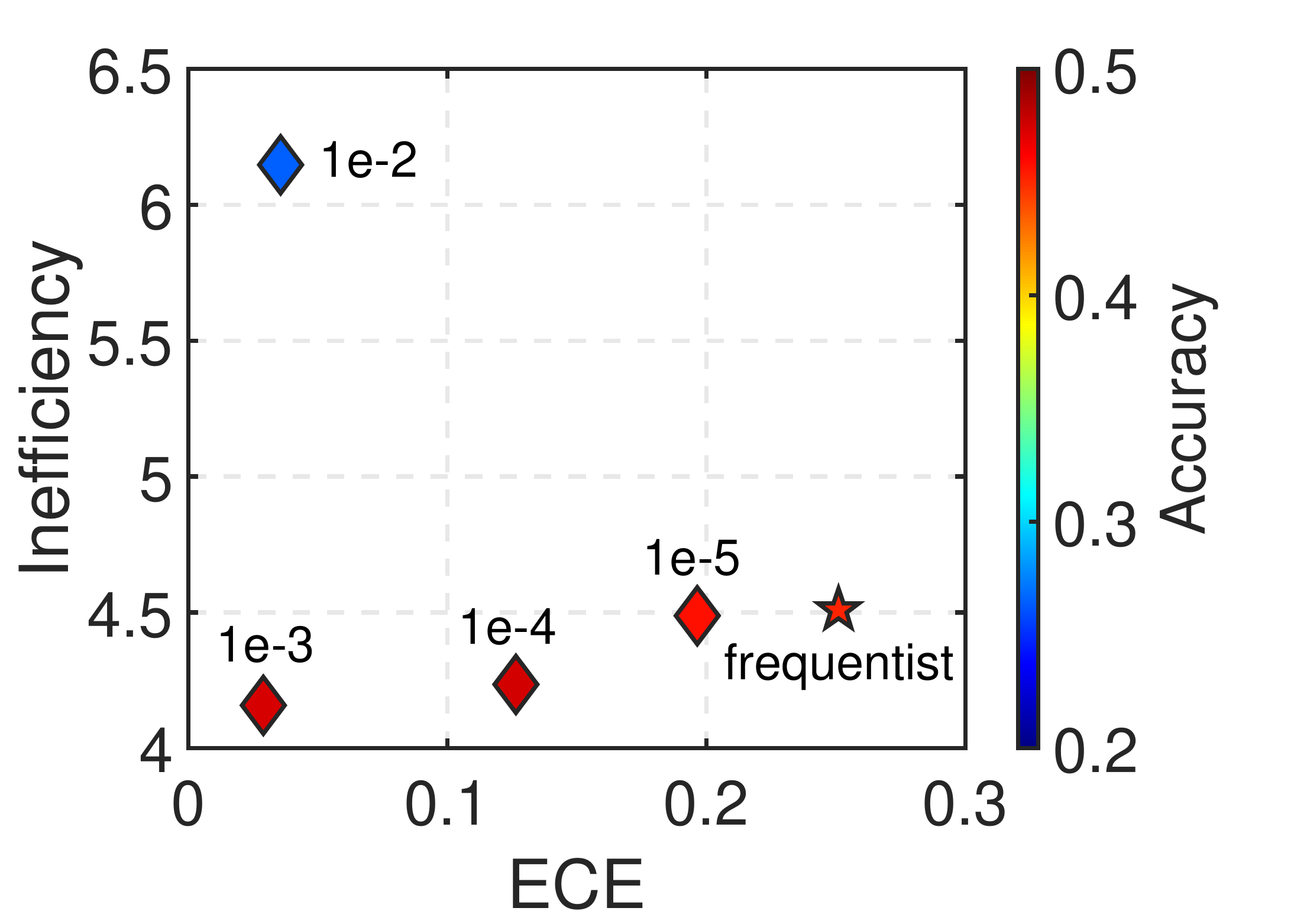}}
        \raisebox{-\height}{\includegraphics[width=0.32\textwidth]{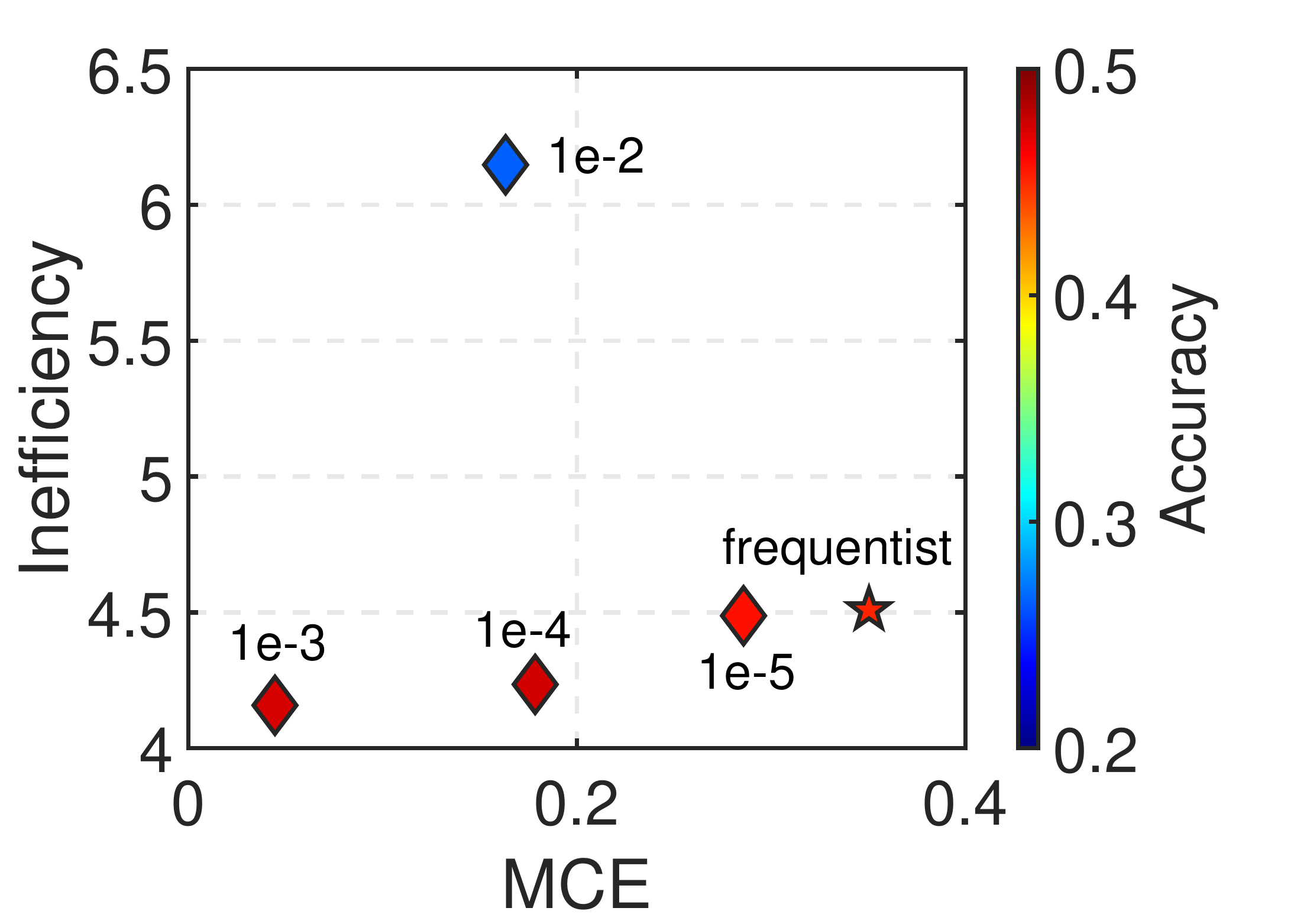}}
        \raisebox{-\height}{\includegraphics[width=0.32\textwidth]{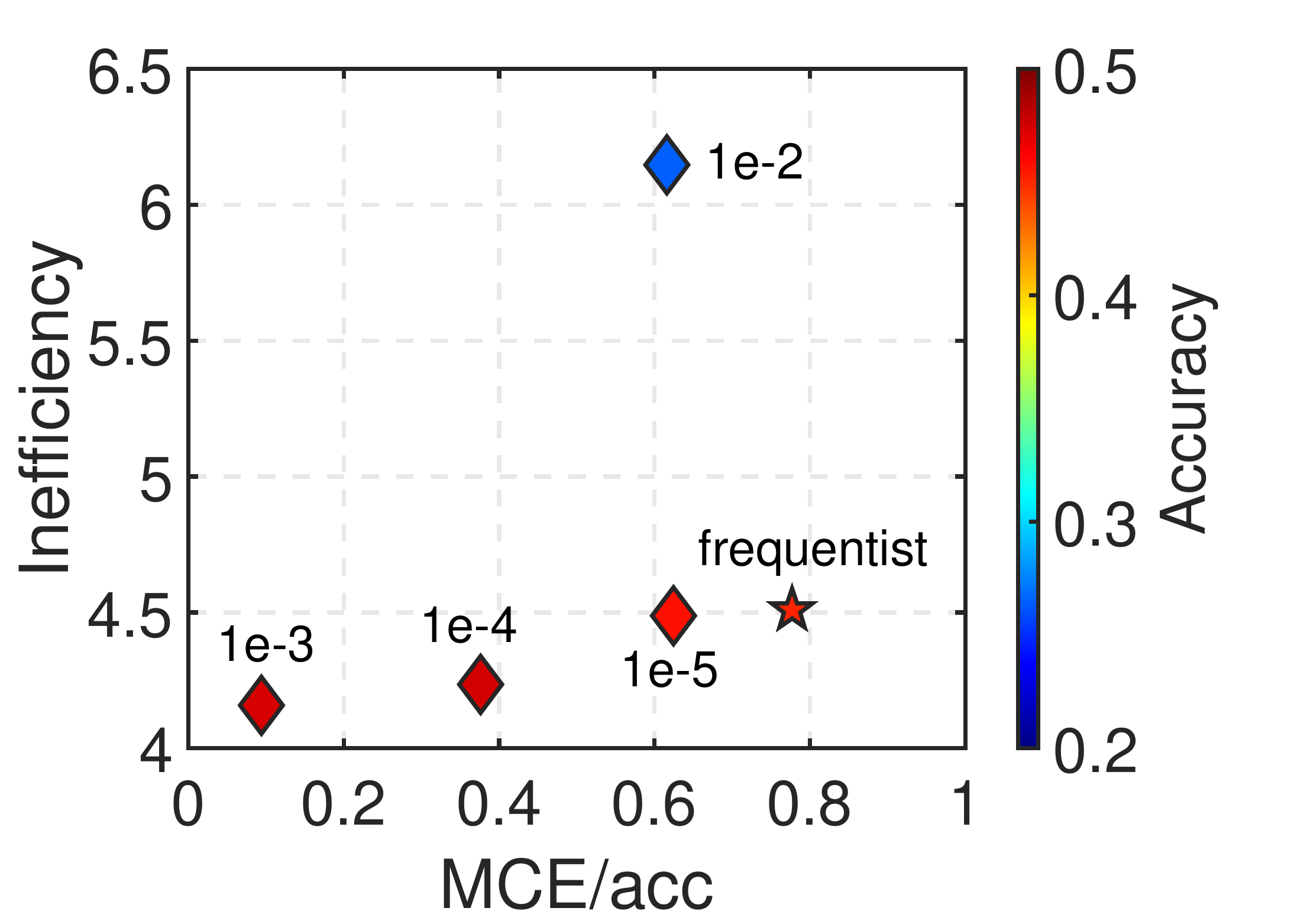}}
        \caption{Relation between inefficiency and calibration measures}
        \vspace{0.3cm}
        \label{subfig:cifar-a}
    \end{subfigure}
    \begin{subfigure}[t]{1\textwidth}
        \raisebox{-\height}{\includegraphics[width=0.32\textwidth]{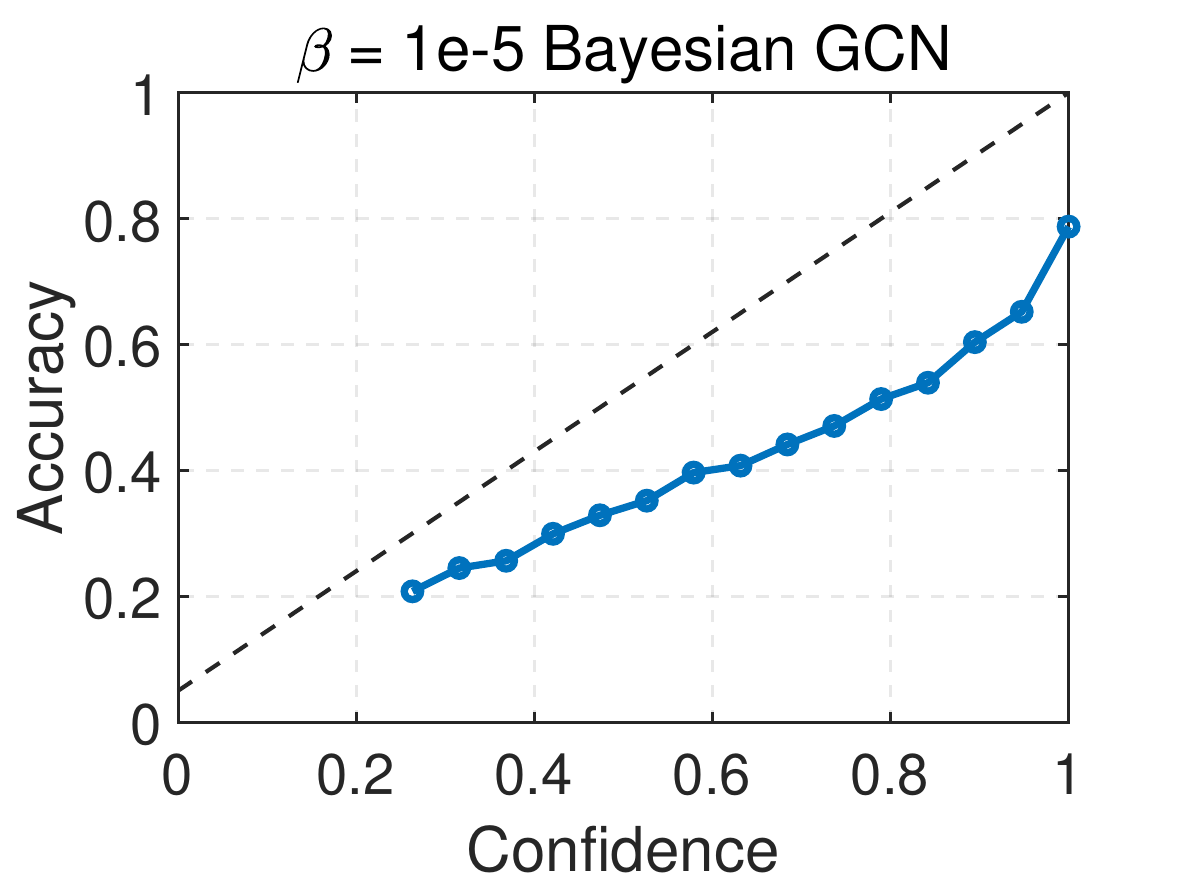}}
        \raisebox{-\height}{\includegraphics[width=0.32\textwidth]{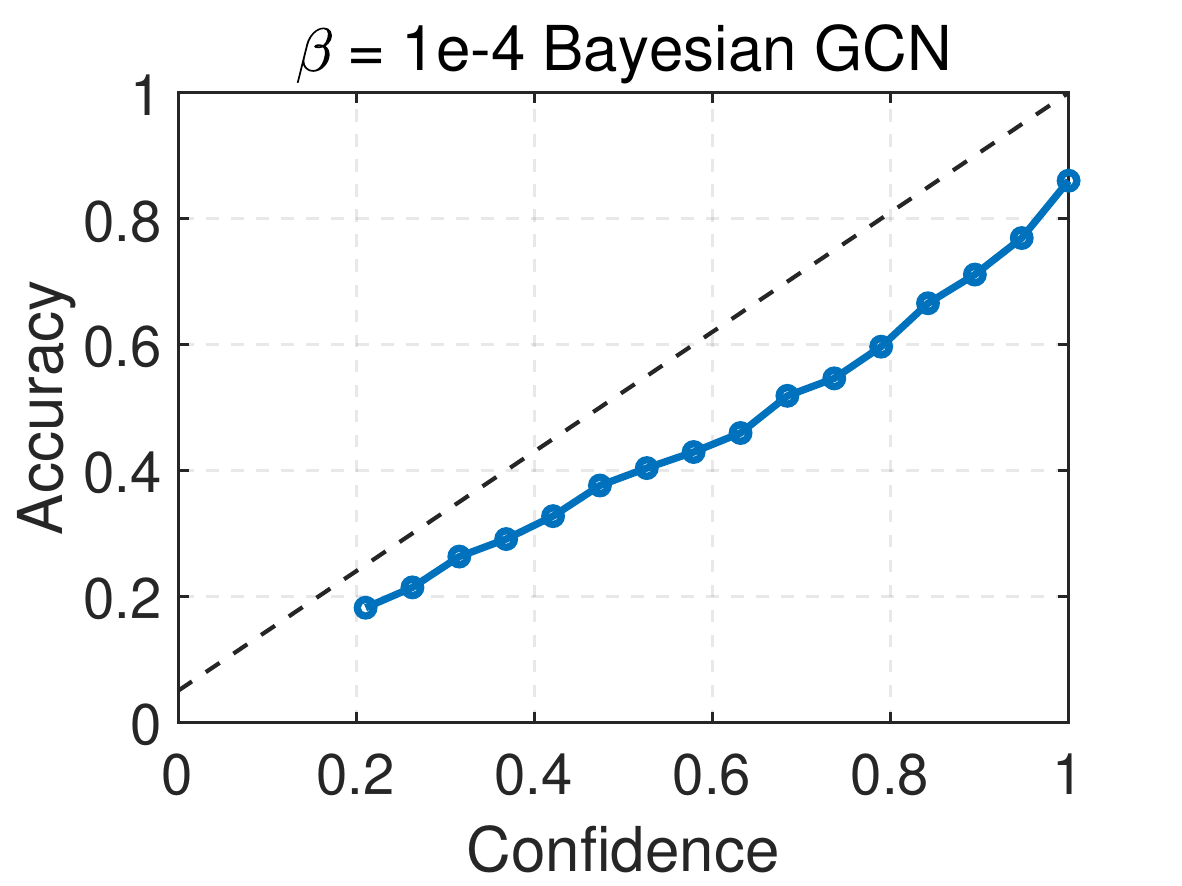}}
        \raisebox{-\height}{\includegraphics[width=0.32\textwidth]{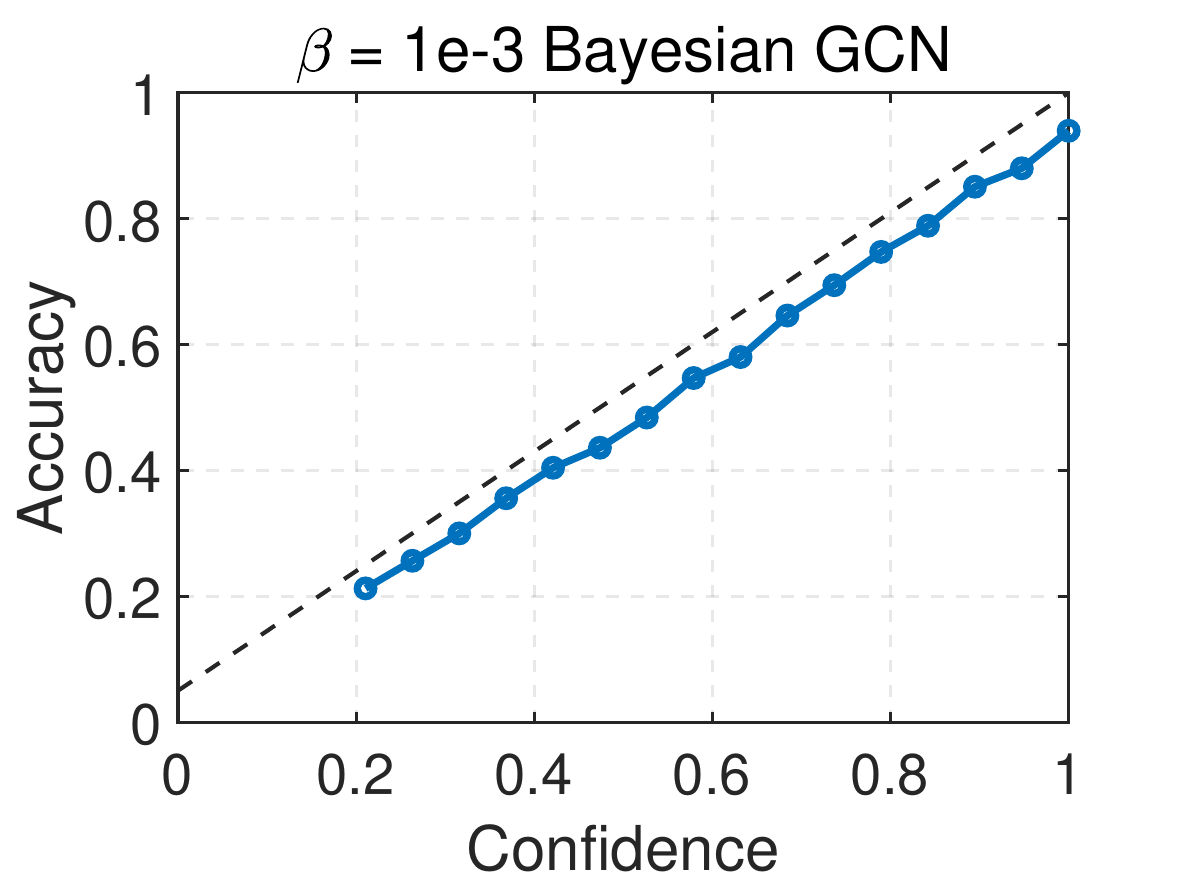}}
        \caption{Reliability diagram of trained Bayesian GCNs on CIFAR10}
        \label{subfig:cifar-b}
    \end{subfigure}
  \caption{\textbf{Results of inefficiency and calibration on CIFAR10}. (a) Model with the lower inefficiency implies that it generates prediction sets with smaller size on average. Bayesian GCNs with $\beta= 10^{-3} $ has the lowest inefficiency. (b) The line $y=x$ on reliability diagrams represents a perfectly calibrated model.}
  \label{fig:cifar10 reliability}
  \vspace{-0.4cm}
\end{figure}

\paragraph{Relation between inefficiency and model calibration} 
To provide a deeper understanding of the connection between prediction sets and the underlying model, we conducted evaluation of Bayesian GCNs with different temperatures, assessing calibration through reliability diagram, ECE and MCE. Overall, we have verified that employing an appropriate temperature parameter $\beta$ enhances the calibration of GCNs. This improvement is evident in reliabillity diagrams, as shown in Fig.~\ref{subfig:cifar-b}. Bayesian GCN model which has the lowest inefficiency at $\beta=10^{-3}$ is comparably well-calibrated than other models.

Furthermore, our experiments imply two major observations. First, \textit{when two models exhibit similar model accuracy levels, the better calibrated model tends to have lower inefficiency, even if the difference can be slight}. In Fig.~\ref{subfig:cite-a}, the Bayesian model with $\beta=4 \cdot10^{-3}$, although having similar accuracy with the model $\beta=8 \cdot10^{-4}$, exhibits reduced inefficiency. This can be attributed to its more robust calibration, as evidenced by ECE, MCE, and the reliability diagram. Similar case can be observed from Fig.~\ref{fig:cifar10 reliability} between two Bayesian models with $\beta=10^{-3}$ and $\beta=10^{-4}$.

The second finding is that \textit{a well-calibrated model, even when its accuracy is relatively low, yields reasonably efficient prediction sets}. From the comparison of Bayesian models with $\beta=2\cdot10^{-3}$ and $\beta=8\cdot10^{-3}$, we clearly see that the well-calibrated one with  $\beta=2\cdot10^{-3}$ results in lower inefficiency, despite its lower test accuracy. Note that the case of $\beta=2\cdot10^{-3}$, the two calibration measures ECE and MCE well reflect the reliability diagram as seen in Fig.~\ref{subfig:cite-a}. However, in the case of a model with $\beta=10^{-2}$, while MCE reasonably captures the low inefficiency resulting from its poor model quality, whereas ECE does not. Hence, in the following, we explore how previous calibration measures, ECE and MCE, reflect CP inefficiency and propose insights on temperature selection through a new calibration measure.

\paragraph{Combined measure for temperature selection} While our previous findings have indicated the existence of temperatures that lead to efficient prediction sets through CP, it is more preferable to find such good temperatures that yield low inefficiency, without adding computational burden from CP. For this purpose, MCE proves to be better than ECE, and this can be attributed to the sensitivity of inefficiency to difficult samples when making prediction sets. However, it is important to note that neither ECE nor MCE perfectly reflects inefficiency, making points such as $\beta=10^{-2}$ in Fig.~\ref{fig:cite reliability} and $\beta=10^{-2}$ in Fig.~\ref{fig:cifar10 reliability}. Consequently, recognizing that both model performance and calibration contribute to set size, we propose that a combined measure involving accuracy and calibration measure could offer a better chance to capture the inefficiency of models with different temperatures. In the rightmost figures of Fig.~\ref{fig:cite reliability} and Fig.~\ref{fig:cifar10 reliability}, we present the relation between the combined measure, calculated as MCE divided by accuracy, and inefficiency. These figures show that the combined measure better accounts for the inefficiency of models with high temperatures, addressing the limitations of ECE and MCE. While we have provided a preliminary exploration of the combined measure's ability to represent the set size in CP, we plan to further investigate a measure that well explains the inefficiency. Additionally, in our future research, we aim to develop methods for optimizing the temperature through CP-aware loss \cite{huang2023uncertainty, yang2021finite, stutz2021learning, park2022few} during training process. 
\vspace{-0.1cm}

\section{Conclusion}
\vspace{-0.1cm}
In this work, we explore the impact of temperature on CP in Bayesian GNNs, specifically focusing on the inefficiency of prediction sets. We show that Bayesian GNNs improve the performance in CP compared to frequentist GNNs while providing more flexibility to control the inefficiency via temperature. Our experiments demonstrate that there exist temperatures of Bayesian GNNs that generate efficient prediction sets within CP. Moreover, our analysis on the connection between inefficiency and model calibration suggests further possibility for a measure capable of accurately capturing the inefficiency. Assessing the CP performance using such measure can reduce the computational burden induced by CP. Furthermore, the investigation of a novel method for training the temperature parameter using CP-aware loss \cite{huang2023uncertainty, yang2021finite, stutz2021learning, park2022few} remains as an interesting future work. We anticipate that these findings can provide guidance to trustworthy graph learning that studies reliable uncertainty estimates employing CP. 

\section*{Acknowledgement}
\vspace{-0.1cm}
The authors want to thank Sangwoo Park and Osvaldo Simeone for providing the main idea of this work that studies the impact of temperature in Bayesian learning when used in conjunction with CP under GNN architecture. This research was supported by the MSIT (Ministry of Science and ICT), Korea, under the ITRC (Information Technology Research Center) support program (IITP-2020-0-01787 and IITP-2023-RS-2023-00259991) supervised by the IITP (Institute of Information \& Communications Technology Planning \& Evaluation).

\newpage
\bibliography{references}

\begin{thebibliography}{10}\itemsep=0pt

\bibitem{abdar2021review}
Moloud Abdar, Farhad Pourpanah, Sadiq Hussain, Dana Rezazadegan, Li Liu, Mohammad Ghavamzadeh, Paul Fieguth, Xiaochun Cao, Abbas Khosravi, U~Rajendra Acharya, et~al.
\newblock A review of uncertainty quantification in deep learning: Techniques, applications and challenges.
\newblock {\em Information fusion}, 76:243--297, 2021.

\bibitem{barber2023conformal}
Rina~Foygel Barber, Emmanuel~J Candes, Aaditya Ramdas, and Ryan~J Tibshirani.
\newblock Conformal prediction beyond exchangeability.
\newblock {\em The Annals of Statistics}, 51(2):816--845, 2023.

\bibitem{blundell2015weight}
Charles Blundell, Julien Cornebise, Koray Kavukcuoglu, and Daan Wierstra.
\newblock Weight uncertainty in neural network.
\newblock In {\em International conference on machine learning}, pages 1613--1622. PMLR, 2015.

\bibitem{boluki2020learnable}
Shahin Boluki, Randy Ardywibowo, Siamak~Zamani Dadaneh, Mingyuan Zhou, and Xiaoning Qian.
\newblock Learnable bernoulli dropout for bayesian deep learning.
\newblock In {\em International Conference on Artificial Intelligence and Statistics}, pages 3905--3916. PMLR, 2020.

\bibitem{chen2018fastgcn}
Jie Chen, Tengfei Ma, and Cao Xiao.
\newblock Fastgcn: fast learning with graph convolutional networks via importance sampling.
\newblock {\em arXiv preprint arXiv:1801.10247}, 2018.

\bibitem{clarkson2023distribution}
Jase Clarkson.
\newblock Distribution free prediction sets for node classification.
\newblock In {\em International Conference on Machine Learning}, pages 6268--6278. PMLR, 2023.

\bibitem{dhillon2023expected}
Guneet~S Dhillon, George Deligiannidis, and Tom Rainforth.
\newblock On the expected size of conformal prediction sets.
\newblock {\em arXiv preprint arXiv:2306.07254}, 2023.

\bibitem{dwivedi2020benchmarking}
Vijay~Prakash Dwivedi, Chaitanya~K Joshi, Anh~Tuan Luu, Thomas Laurent, Yoshua Bengio, and Xavier Bresson.
\newblock Benchmarking graph neural networks.
\newblock {\em arXiv preprint arXiv:2003.00982}, 2020.

\bibitem{gal2016dropout}
Yarin Gal and Zoubin Ghahramani.
\newblock Dropout as a bayesian approximation: Representing model uncertainty in deep learning.
\newblock In {\em international conference on machine learning}, pages 1050--1059. PMLR, 2016.

\bibitem{grunwald2017inconsistency}
Peter Gr{\"u}nwald and Thijs Van~Ommen.
\newblock Inconsistency of bayesian inference for misspecified linear models, and a proposal for repairing it.
\newblock 2017.

\bibitem{guo2017calibration}
Chuan Guo, Geoff Pleiss, Yu Sun, and Kilian~Q Weinberger.
\newblock On calibration of modern neural networks.
\newblock In {\em International conference on machine learning}, pages 1321--1330. PMLR, 2017.

\bibitem{hamilton2017inductive}
Will Hamilton, Zhitao Ying, and Jure Leskovec.
\newblock Inductive representation learning on large graphs.
\newblock {\em Advances in neural information processing systems}, 30, 2017.

\bibitem{hasanzadeh2020bayesian}
Arman Hasanzadeh, Ehsan Hajiramezanali, Shahin Boluki, Mingyuan Zhou, Nick Duffield, Krishna Narayanan, and Xiaoning Qian.
\newblock Bayesian graph neural networks with adaptive connection sampling.
\newblock In {\em International conference on machine learning}, pages 4094--4104. PMLR, 2020.

\bibitem{hsu2022makes}
Hans Hao-Hsun Hsu, Yuesong Shen, Christian Tomani, and Daniel Cremers.
\newblock What makes graph neural networks miscalibrated?
\newblock {\em Advances in Neural Information Processing Systems}, 35:13775--13786, 2022.

\bibitem{huang2023uncertainty}
Kexin Huang, Ying Jin, Emmanuel Candes, and Jure Leskovec.
\newblock Uncertainty quantification over graph with conformalized graph neural networks.
\newblock {\em arXiv preprint arXiv:2305.14535}, 2023.

\bibitem{hullermeier2021aleatoric}
Eyke H{\"u}llermeier and Willem Waegeman.
\newblock Aleatoric and epistemic uncertainty in machine learning: An introduction to concepts and methods.
\newblock {\em Machine Learning}, 110:457--506, 2021.

\bibitem{jose2021free}
Sharu~Theresa Jose and Osvaldo Simeone.
\newblock Free energy minimization: A unified framework for modeling, inference, learning, and optimization.
\newblock {\em IEEE Signal Processing Magazine}, 38(2):120--125, 2021.

\bibitem{kipf2016semi}
Thomas~N Kipf and Max Welling.
\newblock Semi-supervised classification with graph convolutional networks.
\newblock {\em arXiv preprint arXiv:1609.02907}, 2016.

\bibitem{knoblauch2019generalized}
Jeremias Knoblauch, Jack Jewson, and Theodoros Damoulas.
\newblock Generalized variational inference: Three arguments for deriving new posteriors.
\newblock {\em arXiv preprint arXiv:1904.02063}, 2019.

\bibitem{knoblauch2022optimization}
Jeremias Knoblauch, Jack Jewson, and Theodoros Damoulas.
\newblock An optimization-centric view on bayes' rule: Reviewing and generalizing variational inference.
\newblock {\em The Journal of Machine Learning Research}, 23(1):5789--5897, 2022.

\bibitem{knyazev2019understanding}
Boris Knyazev, Graham~W Taylor, and Mohamed Amer.
\newblock Understanding attention and generalization in graph neural networks.
\newblock {\em Advances in neural information processing systems}, 32, 2019.

\bibitem{kuchibhotla2020exchangeability}
Arun~Kumar Kuchibhotla.
\newblock Exchangeability, conformal prediction, and rank tests.
\newblock {\em arXiv preprint arXiv:2005.06095}, 2020.

\bibitem{masegosa2020learning}
Andres Masegosa.
\newblock Learning under model misspecification: Applications to variational and ensemble methods.
\newblock {\em Advances in Neural Information Processing Systems}, 33:5479--5491, 2020.

\bibitem{morris2020tudataset}
Christopher Morris, Nils~M Kriege, Franka Bause, Kristian Kersting, Petra Mutzel, and Marion Neumann.
\newblock Tudataset: A collection of benchmark datasets for learning with graphs.
\newblock {\em arXiv preprint arXiv:2007.08663}, 2020.

\bibitem{murphy2012machine}
Kevin~P Murphy.
\newblock {\em Machine learning: a probabilistic perspective}.
\newblock MIT press, 2012.

\bibitem{park2022few}
Sangwoo Park, Kfir~M Cohen, and Osvaldo Simeone.
\newblock Few-shot calibration of set predictors via meta-learned cross-validation-based conformal prediction.
\newblock {\em arXiv preprint arXiv:2210.03067}, 2022.

\bibitem{rong2019dropedge}
Yu Rong, Wenbing Huang, Tingyang Xu, and Junzhou Huang.
\newblock Dropedge: Towards deep graph convolutional networks on node classification.
\newblock {\em arXiv preprint arXiv:1907.10903}, 2019.

\bibitem{sen2008collective}
Prithviraj Sen, Galileo Namata, Mustafa Bilgic, Lise Getoor, Brian Galligher, and Tina Eliassi-Rad.
\newblock Collective classification in network data.
\newblock {\em AI magazine}, 29(3):93--93, 2008.

\bibitem{simeone2022machine}
Osvaldo Simeone.
\newblock {\em Machine learning for engineers}.
\newblock Cambridge university press, 2022.

\bibitem{srivastava2014dropout}
Nitish Srivastava, Geoffrey Hinton, Alex Krizhevsky, Ilya Sutskever, and Ruslan Salakhutdinov.
\newblock Dropout: a simple way to prevent neural networks from overfitting.
\newblock {\em The journal of machine learning research}, 15(1):1929--1958, 2014.

\bibitem{stadler2021graph}
Maximilian Stadler, Bertrand Charpentier, Simon Geisler, Daniel Z{\"u}gner, and Stephan G{\"u}nnemann.
\newblock Graph posterior network: Bayesian predictive uncertainty for node classification.
\newblock {\em Advances in Neural Information Processing Systems}, 34:18033--18048, 2021.

\bibitem{stutz2021learning}
David Stutz, Ali~Taylan Cemgil, Arnaud Doucet, et~al.
\newblock Learning optimal conformal classifiers.
\newblock {\em arXiv preprint arXiv:2110.09192}, 2021.

\bibitem{syring2019calibrating}
Nicholas Syring and Ryan Martin.
\newblock Calibrating general posterior credible regions.
\newblock {\em Biometrika}, 106(2):479--486, 2019.

\bibitem{teixeira2019graph}
Leonardo Teixeira, Brian Jalaian, and Bruno Ribeiro.
\newblock Are graph neural networks miscalibrated?
\newblock {\em arXiv preprint arXiv:1905.02296}, 2019.

\bibitem{tibshirani2019conformal}
Ryan~J Tibshirani, Rina Foygel~Barber, Emmanuel Candes, and Aaditya Ramdas.
\newblock Conformal prediction under covariate shift.
\newblock {\em Advances in neural information processing systems}, 32, 2019.

\bibitem{vovk2005algorithmic}
Vladimir Vovk, Alexander Gammerman, and Glenn Shafer.
\newblock {\em Algorithmic learning in a random world}, volume~29.
\newblock Springer, 2005.

\bibitem{wang2021confident}
Xiao Wang, Hongrui Liu, Chuan Shi, and Cheng Yang.
\newblock Be confident! towards trustworthy graph neural networks via confidence calibration.
\newblock {\em Advances in Neural Information Processing Systems}, 34:23768--23779, 2021.

\bibitem{wilson2020bayesian}
Andrew~G Wilson and Pavel Izmailov.
\newblock Bayesian deep learning and a probabilistic perspective of generalization.
\newblock {\em Advances in neural information processing systems}, 33:4697--4708, 2020.

\bibitem{xu2018powerful}
Keyulu Xu, Weihua Hu, Jure Leskovec, and Stefanie Jegelka.
\newblock How powerful are graph neural networks?
\newblock {\em arXiv preprint arXiv:1810.00826}, 2018.

\bibitem{yang2021finite}
Yachong Yang and Arun~Kumar Kuchibhotla.
\newblock Finite-sample efficient conformal prediction.
\newblock {\em arXiv preprint arXiv:2104.13871}, 2021.

\bibitem{zargarbashi2023conformal}
Soroush~H Zargarbashi, Simone Antonelli, and Aleksandar Bojchevski.
\newblock Conformal prediction sets for graph neural networks.
\newblock 2023.

\bibitem{zecchin2022robust}
Matteo Zecchin, Sangwoo Park, Osvaldo Simeone, Marios Kountouris, and David Gesbert.
\newblock Robust pacm: Training ensemble models under model misspecification and outliers.
\newblock {\em arXiv preprint arXiv:2203.01859}, 2022.

\bibitem{zhao2020uncertainty}
Xujiang Zhao, Feng Chen, Shu Hu, and Jin-Hee Cho.
\newblock Uncertainty aware semi-supervised learning on graph data.
\newblock {\em Advances in Neural Information Processing Systems}, 33:12827--12836, 2020.

\end{thebibliography}
\bibliographystyle{ieee_fullname}

\newpage
\appendix

{\Large \textbf{Supplementary Material}}

\section{Exchangeability Conditions for GNNs}
\label{app:exch}

The only requirement to ensure a predefined coverage guarantee for the prediction set of CP is the exchangeability between calibration and test data. This is easily satisfied in graph classification tasks where the graph data is sampled in i.i.d. manner from the same distribution of graphs. However, it requires more careful consideration in node classification tasks. 

Recent studies \cite{huang2023uncertainty, zargarbashi2023conformal} have proposed that for common permutation-invariant GNNs, the exchangeability of non-conformity scores for calibration and test data is satisfied in transductive node classification tasks. To be simple, since the selection of calibration nodes does not change the set of non-conformity scores of entire nodes in transductive settings, non-conformity scores remain invariant to the permutation of calibration and test data. Hence, CP can produce valid prediction sets from GNNs. In contrast, in the inductive setting where GNNs are trained only on the training nodes and make prediction on unseen nodes during test time, the exchangeability is violated. This is because the non-conformity scores for calibration nodes are influenced by changes in the graph. In response, a recent work \cite{clarkson2023distribution} has proposed employing non-exchangeable conformal prediction \cite{barber2023conformal} in inductive setting, but this approach is not applicable to the transductive setting. In our work, we focus on graph classification tasks and transductive node classification tasks, where the exchangeability and validity of CP are satisfied.   

\paragraph{Randomness control for exchangeability of Bayeisan GNNs} 
Especially for Bayesian GNNs trained using VI loss in the equation \eqref{GDC VI loss}, ensuring the exchangeability of non-conformity scores requires further control over the randomness involved in Bayesian inference. When generating a predictive distribution $p_\beta(y|x, q)$ from the tempered posterior with $\beta$ as in the equation \eqref{eq:MCsampling}, we ensure that the same drop rate and corresponding random masks are employed for both  calibration and test data to ensure the exchangeability. 

We sample $T$ sets of drop rates for each layer $l \in \{1, \cdots, L\}$ as $\{\pi_t^{(l)}\}_{t=1}^T$ and generate corresponding random masks $\{\mathbf{Z}_t^{(l)}\}_{t=1}^T$. This process yields a fixed set of random model parameters $\{\mathbf{W}_t^{(l)}\}_{t=1}^T$ for each layer, shared across calibration and test set. We then perform $T$ number of forward passes through the GNNs with a fixed set of random model parameters. The resulting softmax outputs are averaged across $T$ runs to create a single predictive distribution, used to calculate non-conformity scores for both calibration and test data. 

\section{Proofs for Theorem 1, 2}
\label{app:pf}
Assuming a fixed temperature $\beta$ in a Bayesian model, we can unify the proofs for both \autoref{thm1: cov} and \autoref{thm2: gnn_cp_cov} since the validity condition is satisfied for any non-conformity score. Therefore, we present a comprehensive proof of the validity condition as outlined in \autoref{thm1: cov}. The subsequent parts involve demonstrating the exchangeability of non-conformity scores between calibration set and a test data sample, followed by proving coverage guarantees for prediction sets. 

\begin{assumption}
\label{assm 1}
   Given that calibration set $\{z_i\}_{i=1}^n=\{(x_i, y_i)\}_{i=1}^n$ and a test data point $z_{n+1}=(x_{n+1}, y_{n+1})$ are exchangeable, i.e., the joint distribution $p(z_1, \dots, z_n, z_{n+1})$ is permutation invariant with respect to $\{z_1, \dots, z_n, z_{n+1}\}$. Formally, the equality $p(z_1, \dots, z_n, z_{n+1}) = p(z_{\pi(1)}, \dots, z_{\pi(n)}, z_{\pi(n+1)})$ is satisfied for any permutation function $\pi(\cdot)$.   
\end{assumption}

If we achieve the exchangeability as described in \autoref{assm 1}, we can establish that the non-conformity scores $s_i = s(z_i|\theta_\mathcal{D})$ are exchangeable because they are only conditioned on the deterministic parameter $\theta_\mathcal{D}$, i.e., 
\begin{equation}
\begin{split}
    (z_1, \dots, z_n, z_{n+1}) \,{\buildrel d \over =}\, (z_{\pi(1)}, \dots, z_{\pi(n)}, z_{\pi(n+1)}) \\
    \iff (s_1, \dots, s_n, s_{n+1}) \,{\buildrel d \over =}\, (s_{\pi(1)}, \dots, s_{\pi(n)}, s_{\pi(n+1)})
\end{split}
\end{equation} 

\newpage
\begin{lemma}
    \label{lemma1}
    If $s_1, \dots, s_n, s_{n+1}$ are exchangeable random variables, then for any $1-\alpha \in (0, 1)$, then the following inequality holds 
    \begin{equation}
        \mathbb{P}[s_{n+1} \le Q_{\alpha}(\{s_i\}_{i=1}^n)] \ge 1-\alpha
    \end{equation}
    where $Q_{\alpha}(\{s_i\}_{i=1}^n)$ is the $\lceil (1-\alpha)(n+1)\rceil$-th smallest value from the set $\{s_1, \dots, s_n\} \cup \{\infty\}$. 
\end{lemma}
\textit{Proof. } By the exchangeability of random variables \{$s_1, \dots, s_n, s_{n+1}$\} , we have the equality \cite{kuchibhotla2020exchangeability}
\begin{equation}
    \mathbb{P}[s_{n+1} \le Q_\alpha(\{s_i\}_{i=1}^{n+1})] \ge 1-\alpha. 
\end{equation}
Furthermore, we have the following equivalence \cite{tibshirani2019conformal}
\begin{equation}
    s_{n+1} > Q_{\alpha}(\{s_i\}_{i=1}^n) \iff s_{n+1} > Q_{\alpha}(\{s_i\}_{i=1}^{n+1})
\end{equation}
as $s_{n+1}$ cannot be strictly larger than itself or $\infty$. It ensures that the non-conformity score $s_{n+1}$ maintains its position regardless of whether the quantile is considered with or without the test data point.

To sum up, given the exchangeability of nonconformity scores on calibration set and a test data point, along with \autoref{lemma1}, we can derive the inequality that satisfies the validity conditions in both \autoref{thm1: cov} and \autoref{thm2: gnn_cp_cov} as
\begin{equation}
    \mathbb{P}[y_{n+1} \in \mathcal{C}_{n, \alpha}(x_{n+1})] = \mathbb{P}[s(z_{n+1}|\theta_\mathcal{D}) \le Q_{\alpha}(\{s(z_i|\theta_\mathcal{D})\}_{i=1}^n)] \ge 1-\alpha.
\end{equation}

\section{Experimental Details}
\label{app:exp}

\subsection{Model training}

\paragraph{Node classification}  We show results on two real citation graph datasets, \textit{Cora} and \textit{Citeseer} datasets. For Bayesian GCN training, we followed the instructions in \url{https://github.com/armanihm/GDC} \cite{hasanzadeh2020bayesian}. We train for 2500 epochs using Adam optimizer with a learning rate of 0.005. To obtain predictive distribution using Bayesian marginalization in the equation \eqref{eq:MCsampling}, we set the number of runs as $T=20, 12$ for Cora and Citeseer, respectively. For frequentist GCN model, we additionally use weight decay of 0.005 and dropout \cite{srivastava2014dropout}, to reproduce reasonable accuracy though it can be considered as one kind of Bayesian learning. We utilize a 4-layer GCN with hidden layer dimension of 128 following \cite{hasanzadeh2020bayesian} for node classification tasks, which is relatively larger than commonly used 2-layer architecture. However, larger GCNs are prone to over-fitting and over-smoothing issues, hence we demonstrate that not only the Bayesian model effectively addresses these challenges, and also temperature scaling provides even more flexibility in making efficient prediction sets in CP. We focus on the transductive setting, where graph topology and features of every node are given to model during training while labels of test nodes are unobserved.

\paragraph{Graph classification} We use two graph classification datasets, with enough number of labels, \textit{CIFAR10-superpixel} \cite{dwivedi2020benchmarking} and \textit{ENZYMES} \cite{morris2020tudataset}. For CIFAR10-superpixel dataset, we train both BBGDC-GCN and frequentist GCN models for 1000 epochs using Adam optimizer with a learning rate of 0.001. GCN model consists of 5 layers of output feature dimension 180 and averaging readout function to make graph-level representation. We mainly followed the code implementation of \url{https://github.com/graphdeeplearning/benchmarking-gnns} \cite{dwivedi2020benchmarking}. For ENZYMES dataset which can be accessed by TUDataset \cite{morris2020tudataset}, we train for 3000 epochs using Adam optimizer with a learning rate of 0.001. GCN model consists of 3 layers of output feature dimension 128 and sum readout function. To obtain predictive distribution using Bayesian marginalization in the equation \eqref{eq:MCsampling}, we set the number of runs as $T=10$ for both datasets.

\section{Variational Beta-Bernoulli Graph DropConnect (BBGDC)}
\label{app:BBGDC}
To perform variational inference (VI) for GNN weight parameters, GDC random masks, and the drop rate, we have assumed a beta-Bernoulli prior for random masks and replace the Beta prior of drop rate for each layer into Kumaraswamy distribution, following assumptions in \cite{hasanzadeh2020bayesian}. To enable the optimization of drop rates together with weight parameters, we used Augment-REINFORCE-Merge (ARM) estimator which can obtain the gradient of loss with respect to drop rates with two forward passes \cite{boluki2020learnable}. 

\section{Results on Cora and ENZYMES}
We additionally present results of the relation between inefficiency and model calibration on Cora and ENZYMES datasets. Two observations that (i) a well-calibrated model between models with similar accuracy shows better efficiency, and (ii) a model with a relatively low accuracy can achieve improved efficiency if it is well-calibrated, suggested in \ref{results:cal} can be seen from these datasets as well. 

\begin{figure}[!h]
     \centering
    \begin{subfigure}[t]{1\textwidth}
        \raisebox{-\height}{\includegraphics[width=0.32\textwidth]{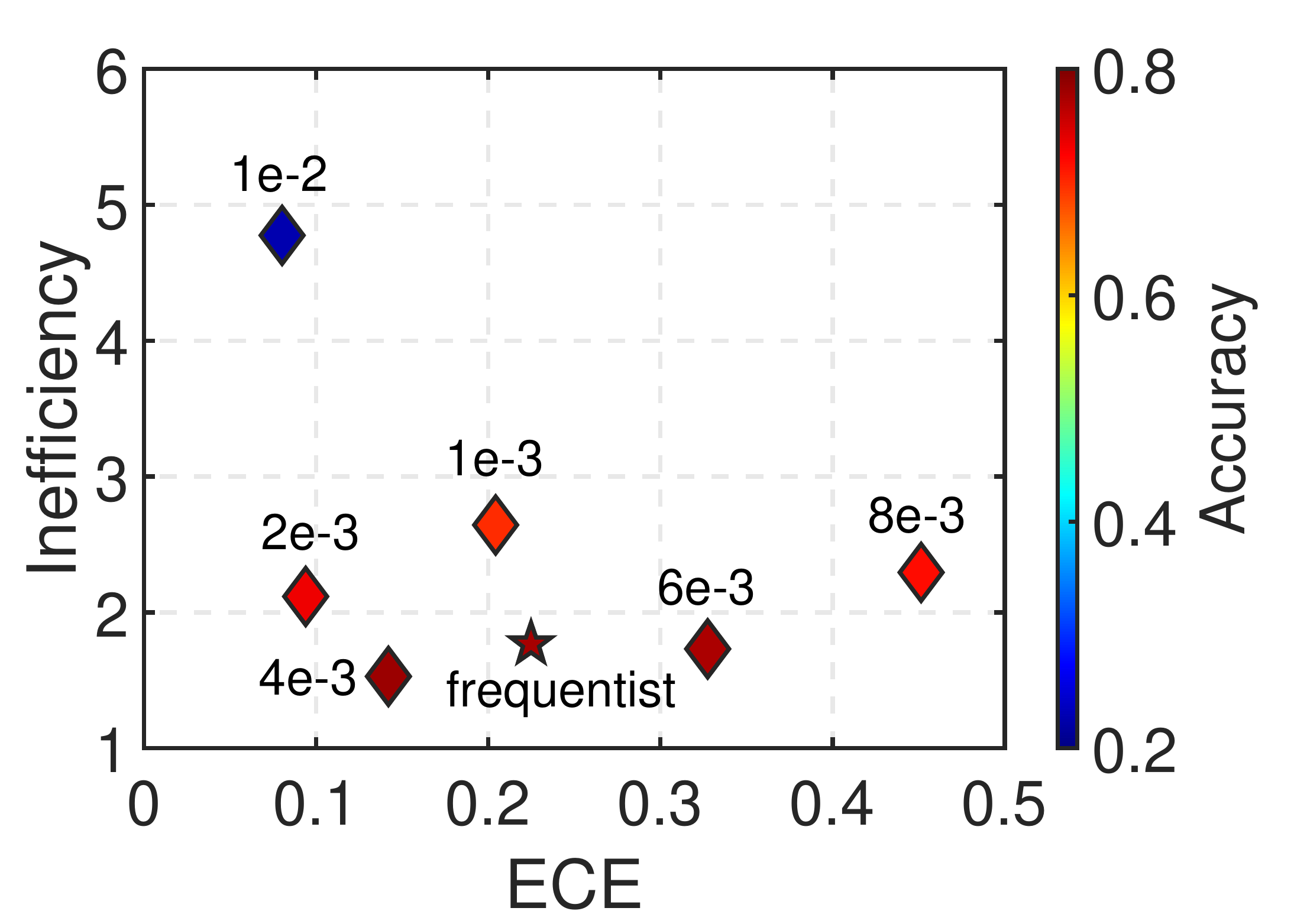}}
        \raisebox{-\height}{\includegraphics[width=0.32\textwidth]{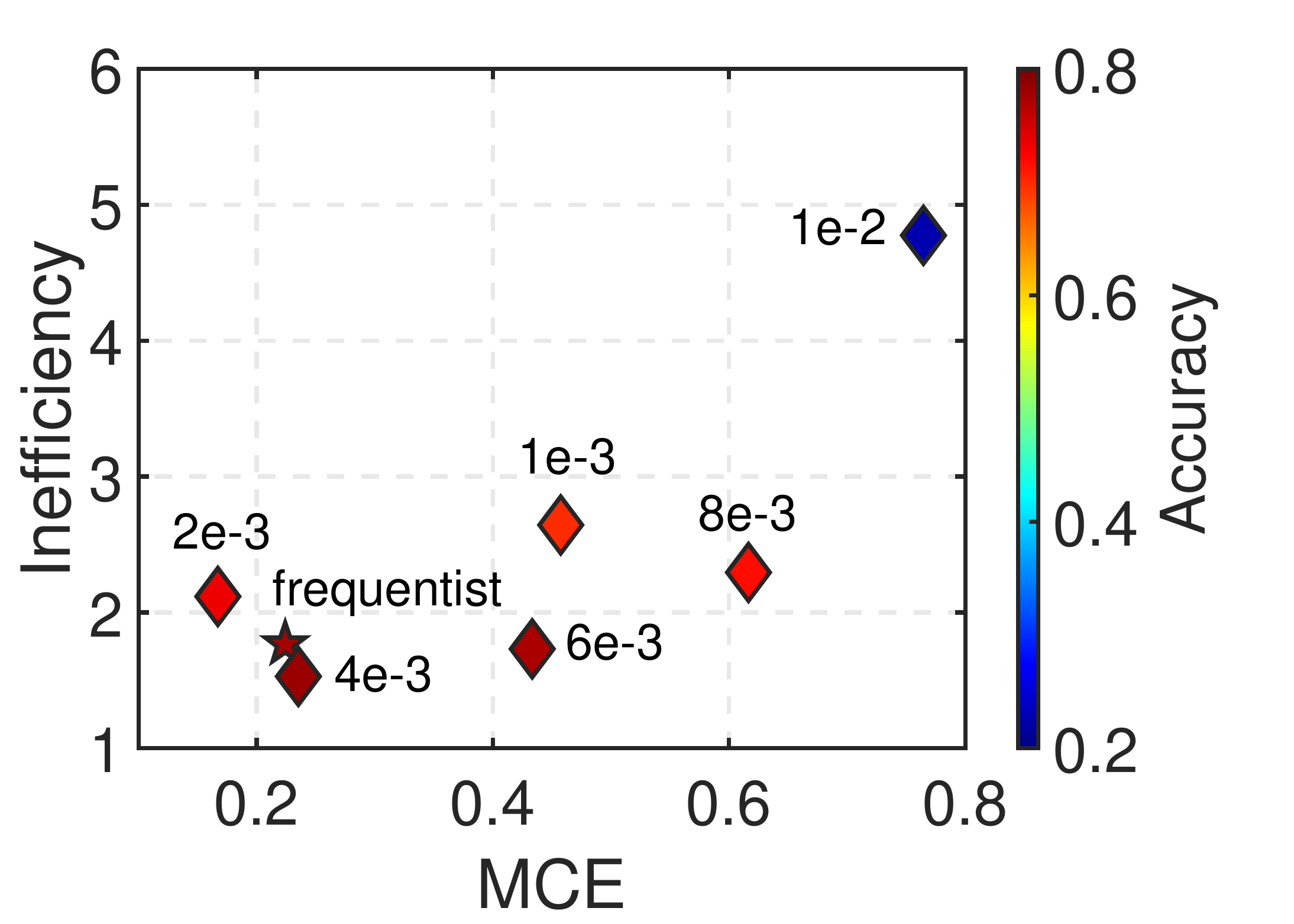}}
        \raisebox{-\height}{\includegraphics[width=0.32\textwidth]{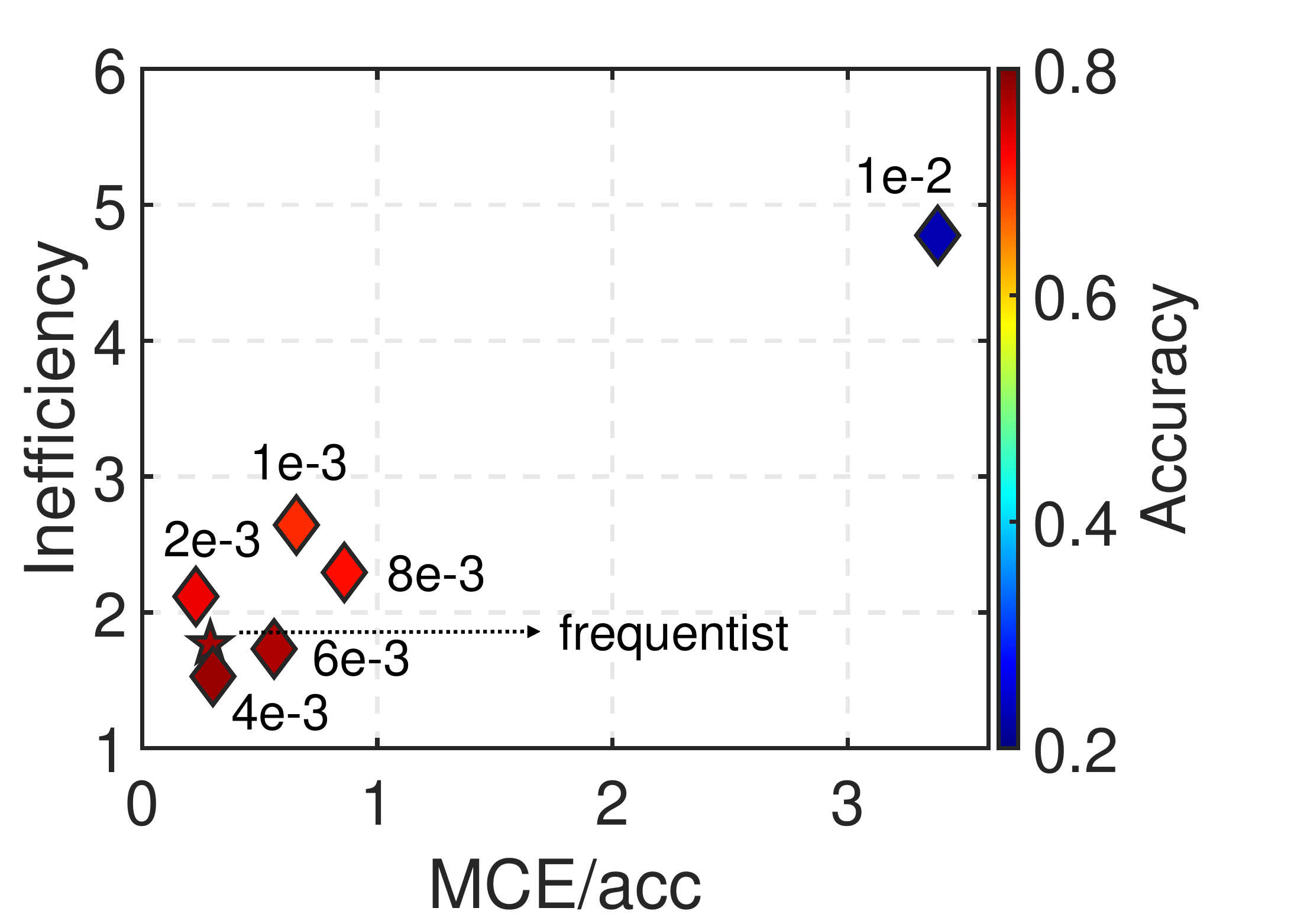}}
        \caption{Relation between inefficiency and calibration measures}
        \label{subfig:cora-a}
        \vspace{0.3cm}
    \end{subfigure}
    \begin{subfigure}[t]{1\textwidth}
        \raisebox{-\height}{\includegraphics[width=0.32\textwidth]{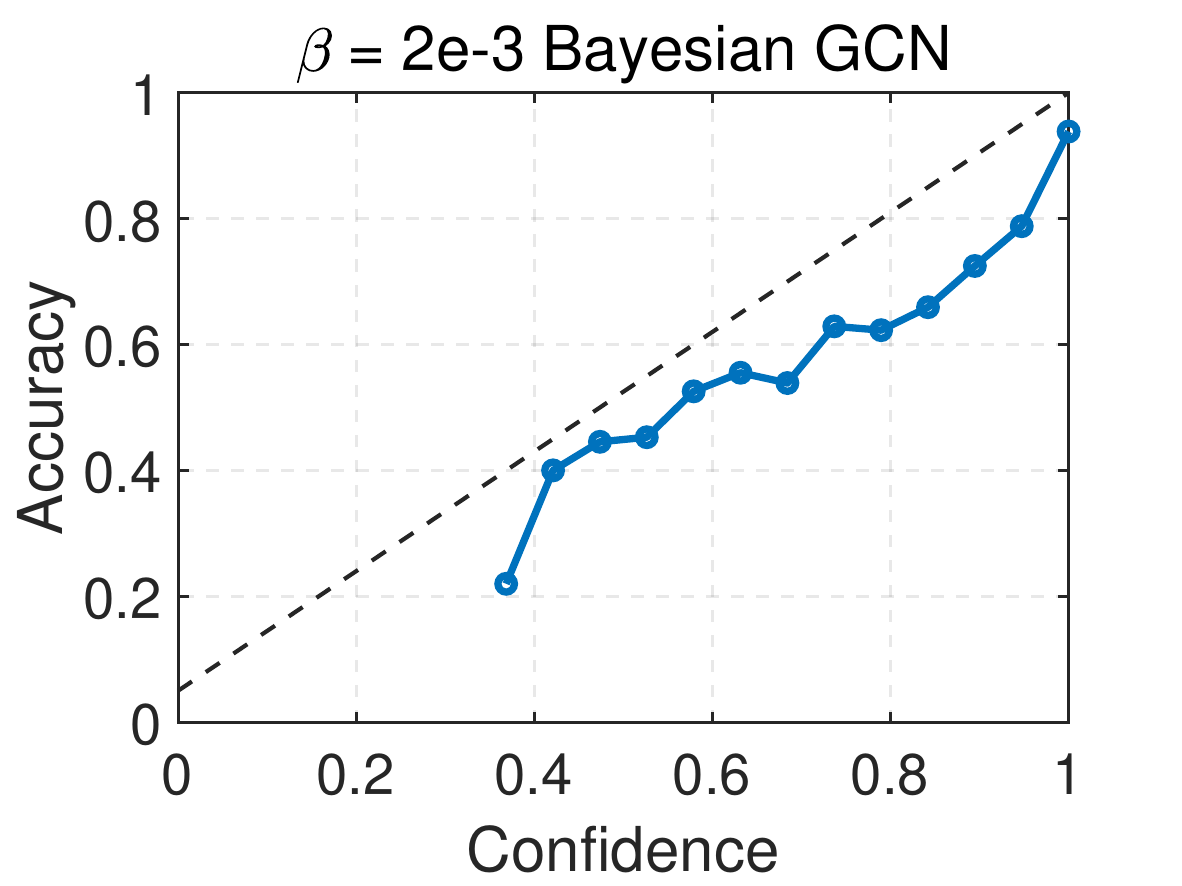}}
        \raisebox{-\height}{\includegraphics[width=0.32\textwidth]{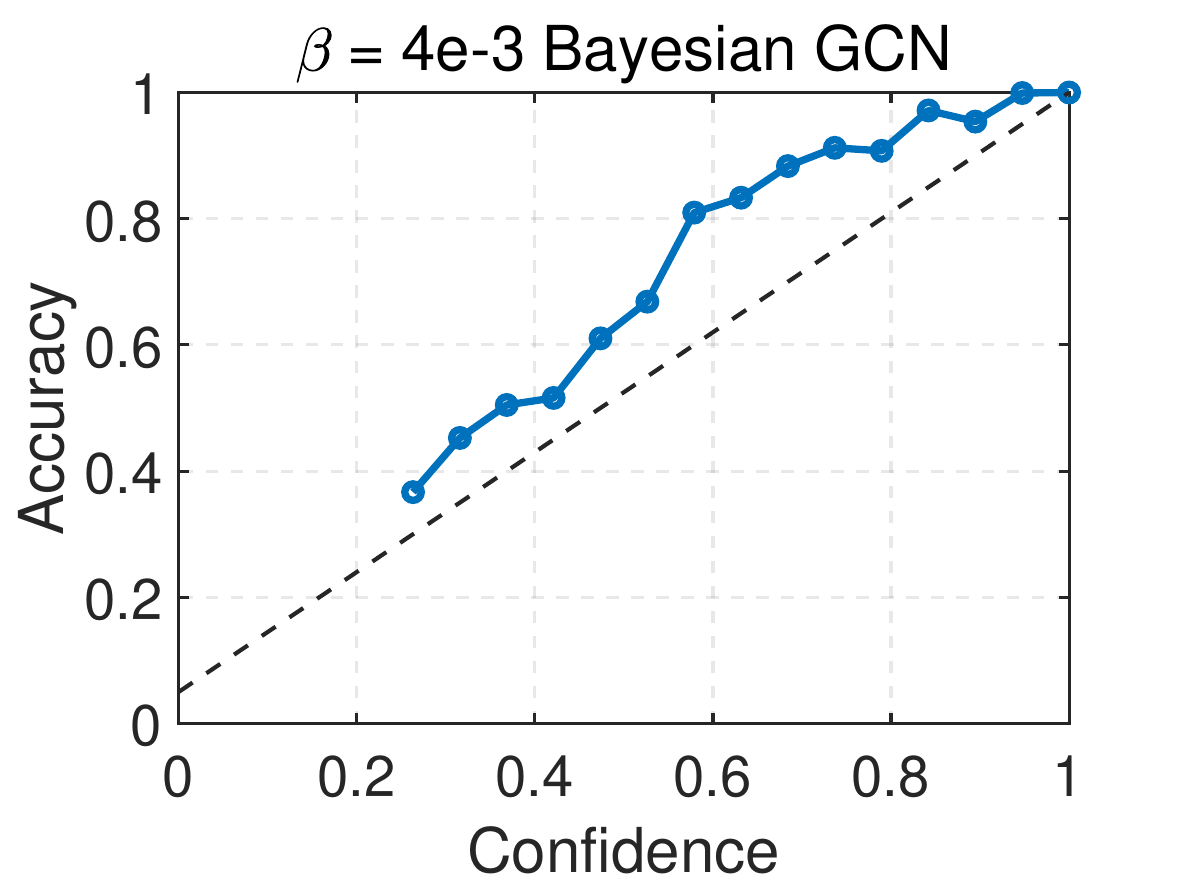}}
        \raisebox{-\height}{\includegraphics[width=0.32\textwidth]{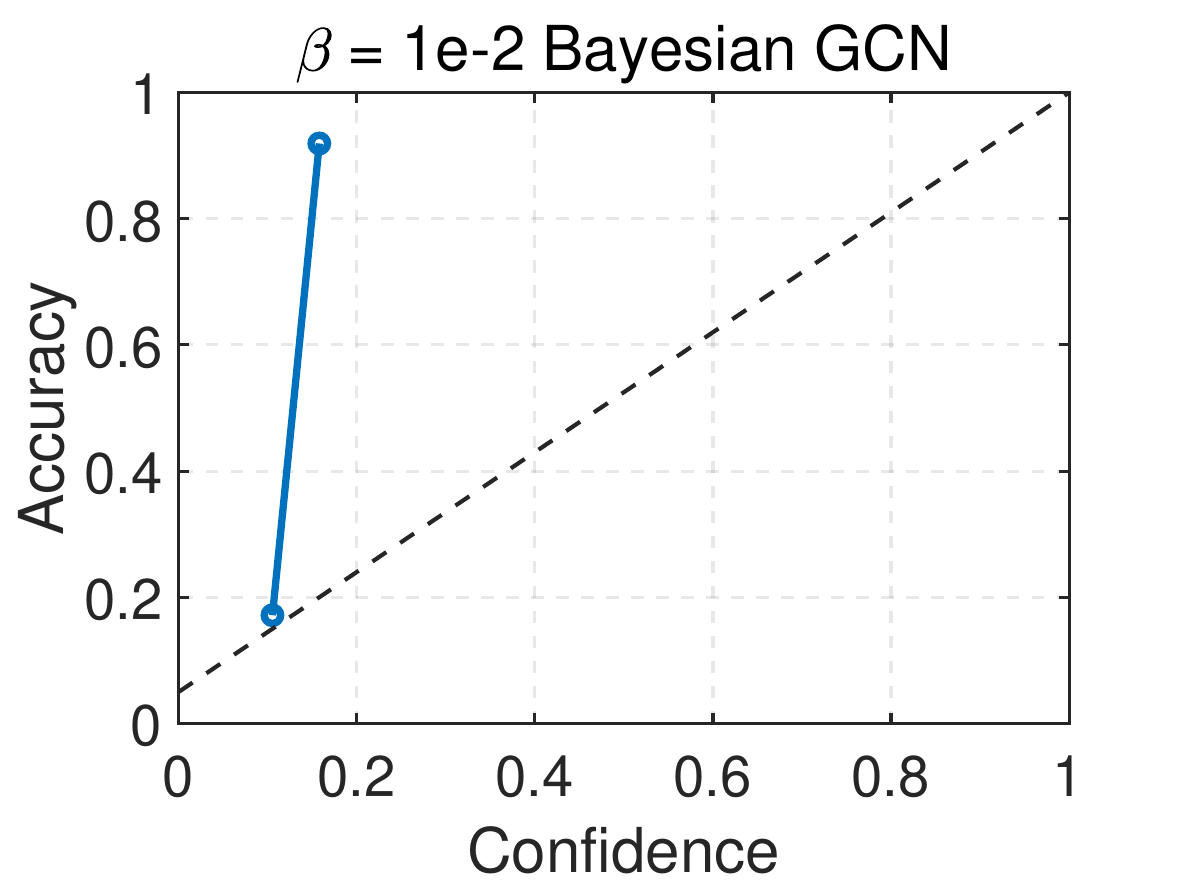}}
        \caption{Reliability diagram of trained GCNs on Cora}
        \label{subfig:cora-b}
    \end{subfigure}
  \caption{\textbf{Results of inefficiency and calibration on Cora}. (a) Model with the lower inefficiency implies that it generates prediction sets with smaller size on average. Bayesian GCNs with $\beta= 4 \cdot10^{-3} $ has the lowest inefficiency. (b) The line $y=x$ on reliability diagrams represents a perfectly calibrated model.}
  \label{fig:cora reliability}
\end{figure}

On Cora dataset, as shown in Fig.~\ref{fig:cora reliability}, among Bayesian GCNs achieving similar test accuracies in the temperature sets $\{10^{-3}, 2\cdot10^{-3}, 8\cdot10^{-3}\}$ and $\{4\cdot10^{-3}, 6\cdot10^{-3}\}$, the better calibrated models, with $\beta = 2\cdot10^{-3}$ and $\beta = 4 \cdot10^{-3}$, in terms of both ECE and MCE exhibit improved efficiency, showing our first observation. In the case of GCNs with $\beta=10^{-2}$ that achieves relatively low test accuracy, it shows the lowest ECE and the highest MCE among models. This is due to the behavior observed in the reliability diagram, where the majority of samples have both low confidence and low accuracy, while a few samples have significantly higher accuracy within a specific bin, leading to a high MCE. This highlights the limitation of ECE to capture the inefficiency in CP and emphasizes the necessity for a new measure. 

\begin{figure}[!h]
     \centering
    \begin{subfigure}[t]{1\textwidth}
        \raisebox{-\height}{\includegraphics[width=0.32\textwidth]{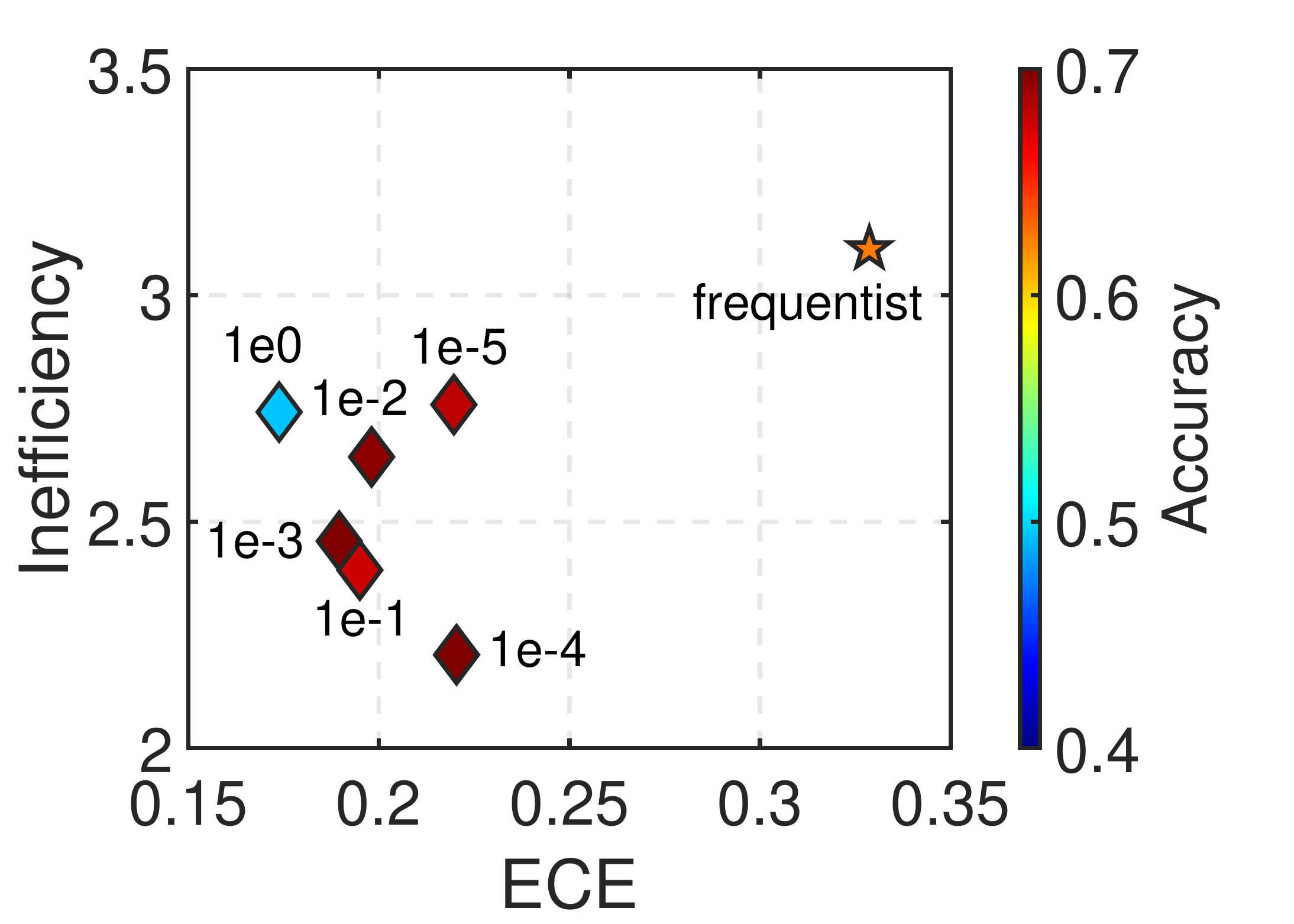}}
        \raisebox{-\height}{\includegraphics[width=0.32\textwidth]{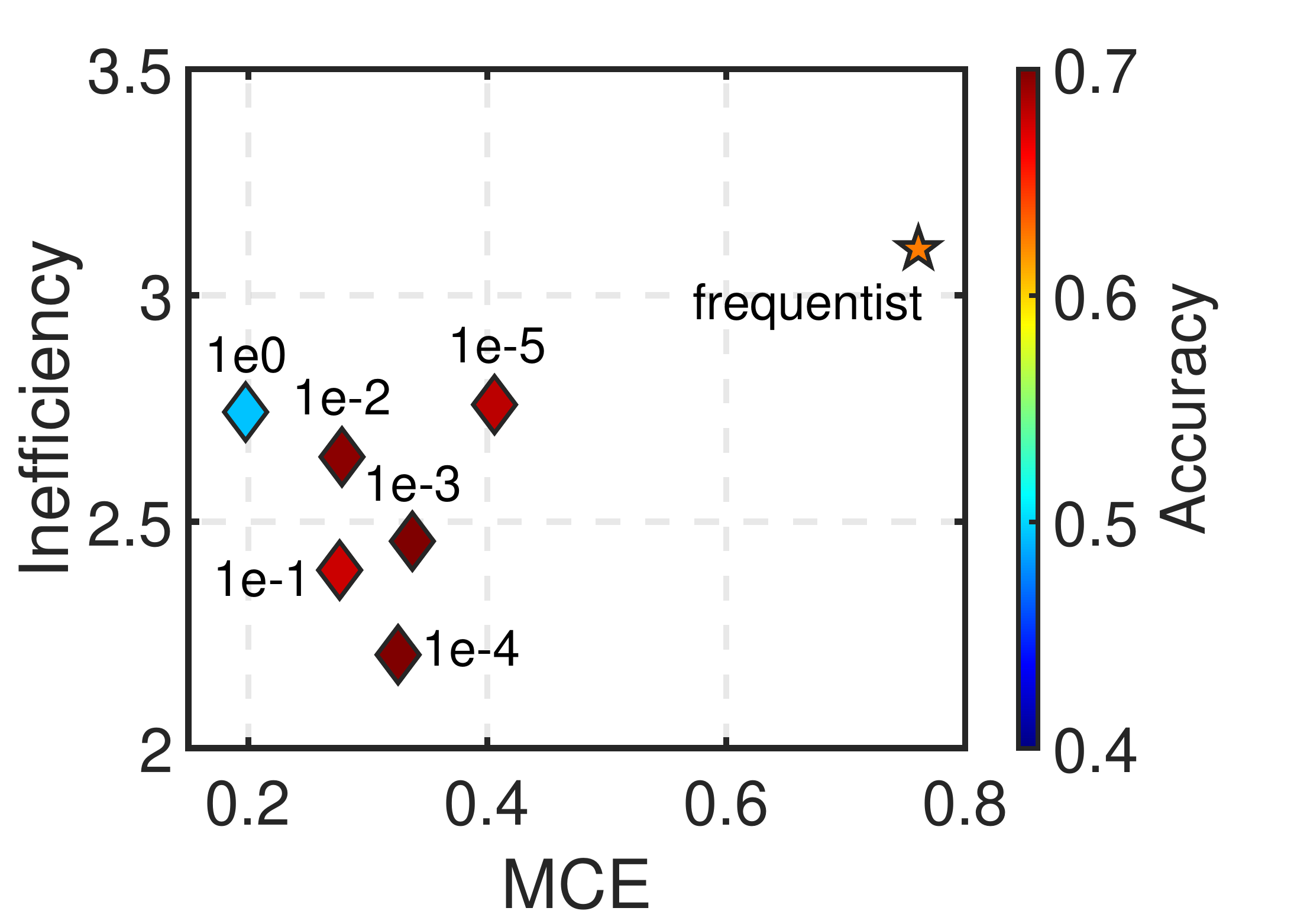}}
        \raisebox{-\height}{\includegraphics[width=0.32\textwidth]{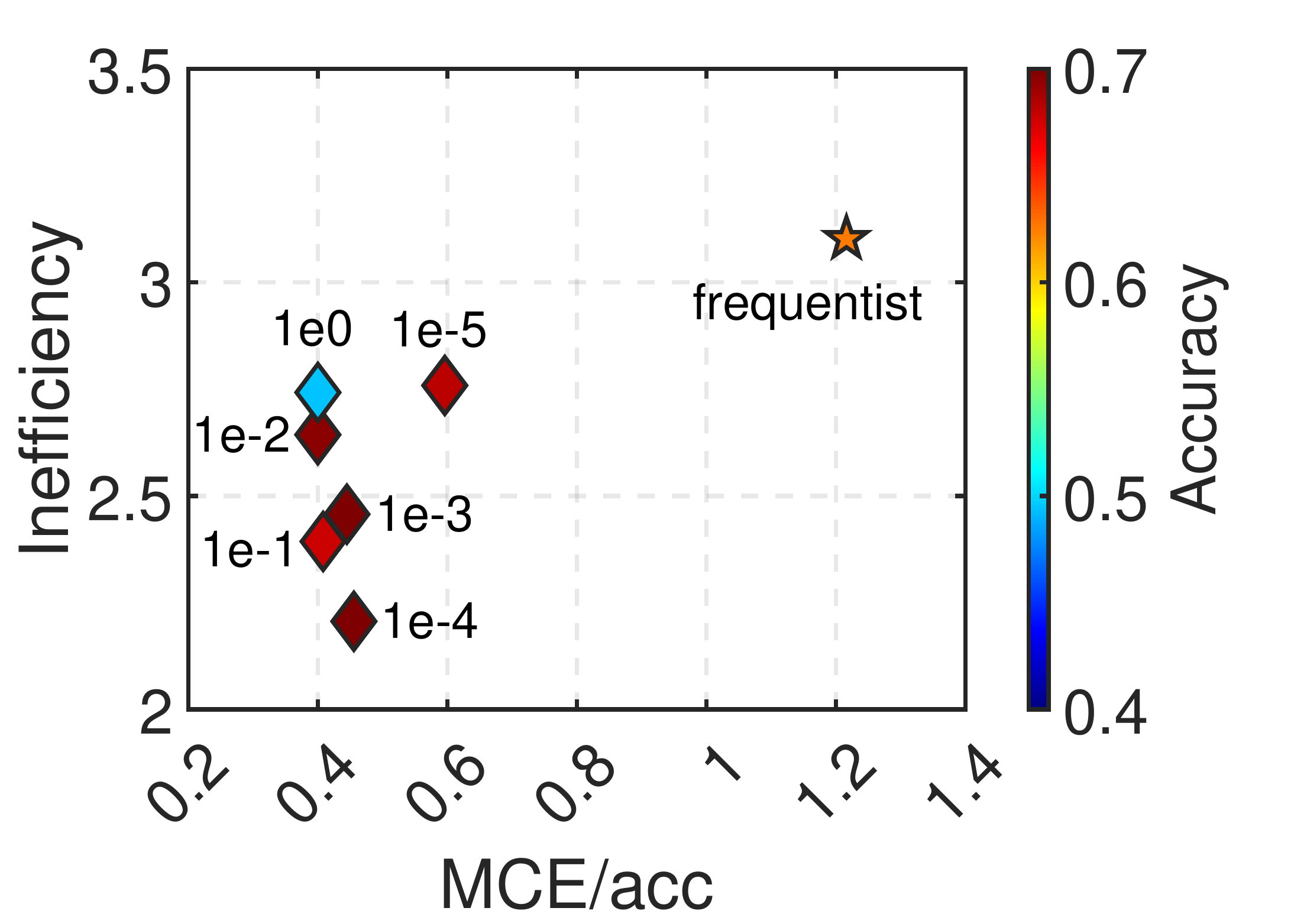}}
        \caption{Relation between inefficiency and calibration measures}
        \label{subfig:enz-a}
        \vspace{0.3cm}
    \end{subfigure}
    \begin{subfigure}[t]{1\textwidth}
        \raisebox{-\height}{\includegraphics[width=0.32\textwidth]{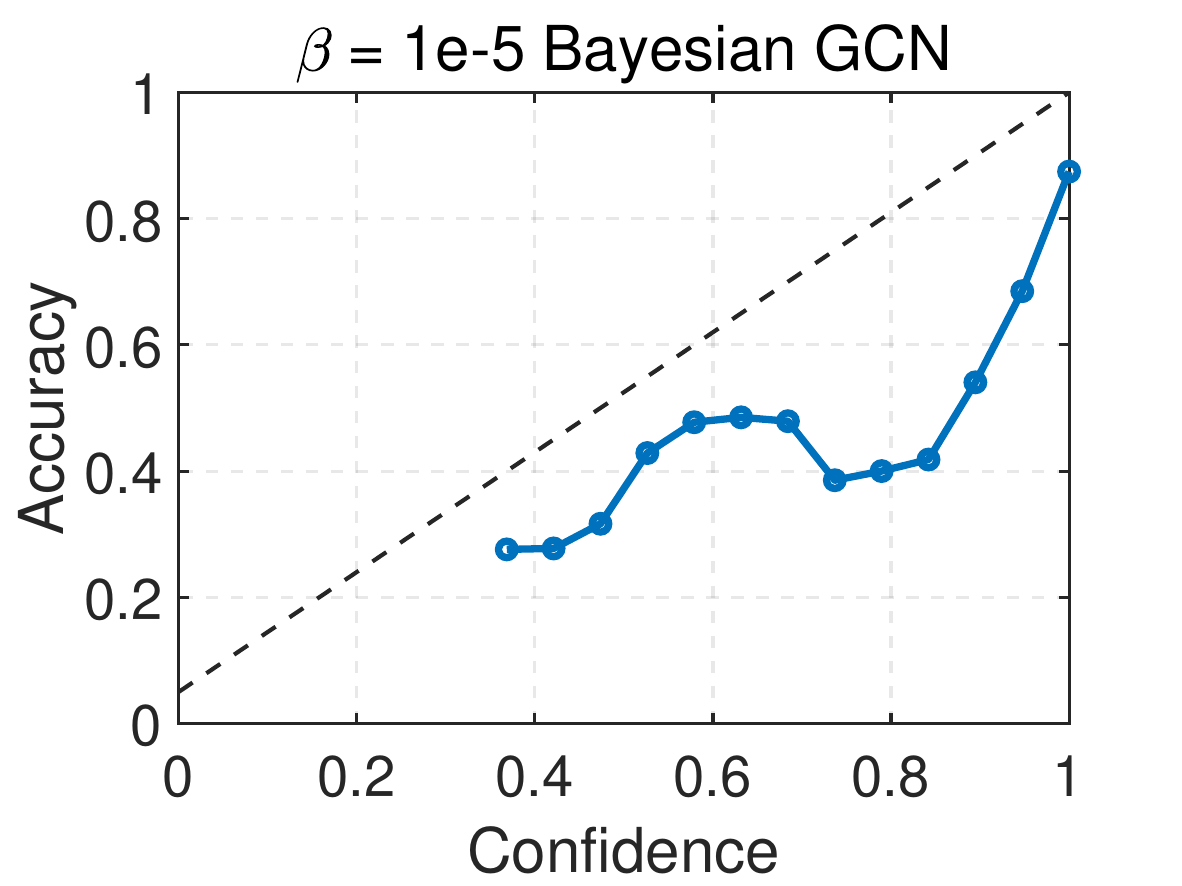}}
        \raisebox{-\height}{\includegraphics[width=0.32\textwidth]{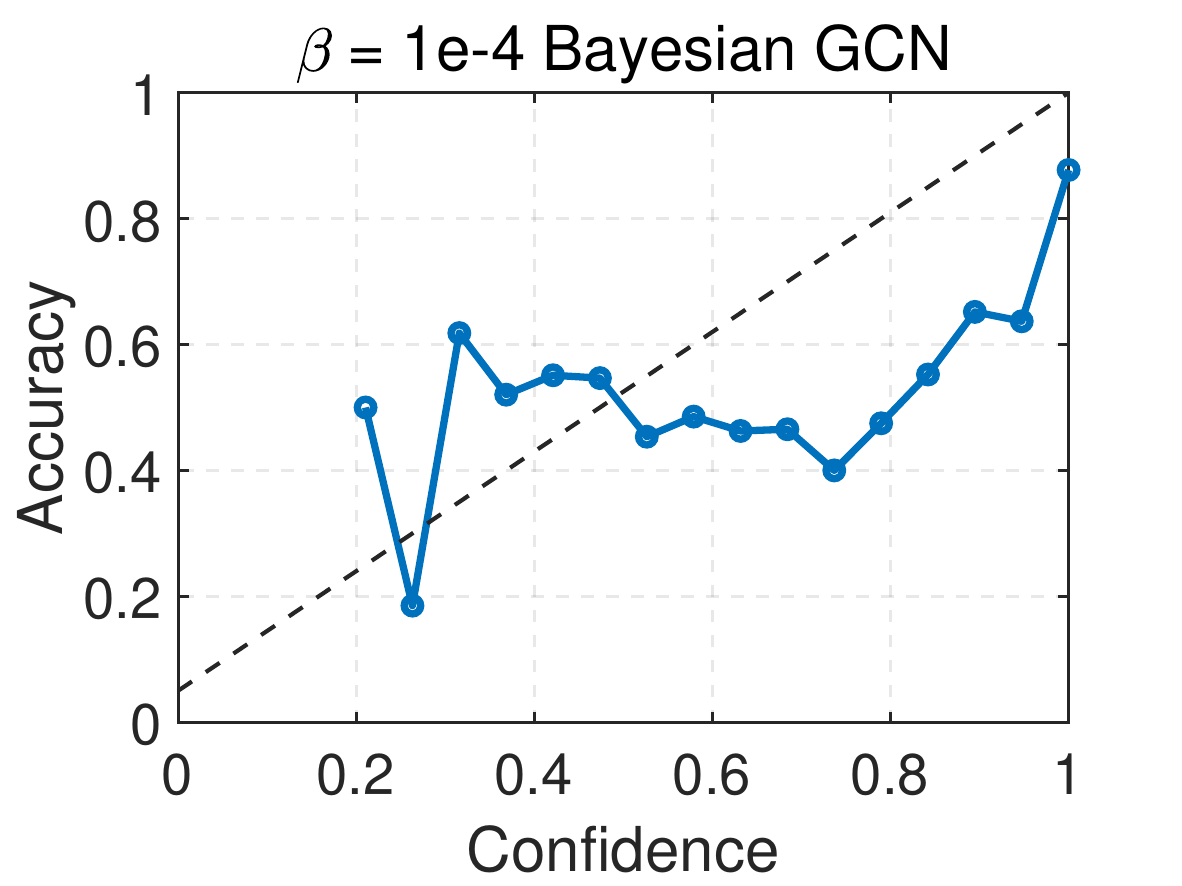}}
        \raisebox{-\height}{\includegraphics[width=0.32\textwidth]{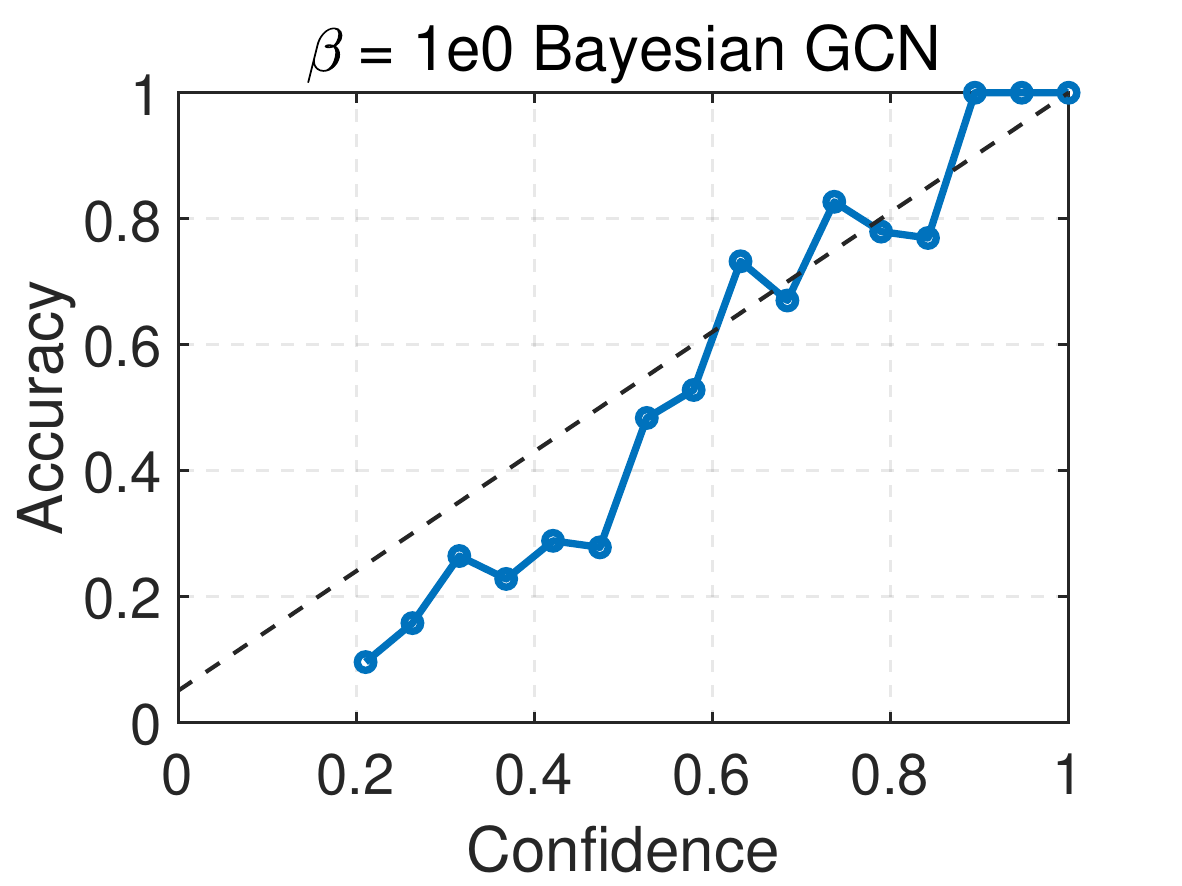}}
        \caption{Reliability diagram of trained GCNs on ENZYMES}
        \label{subfig:enz-b}
    \end{subfigure}
  \caption{\textbf{Results of inefficiency and calibration on ENZYMES}. (a) Model with the lower inefficiency implies that it generates prediction sets with smaller size on average. Bayesian GCNs with $\beta= 10^{-4} $ has the lowest inefficiency. (b) The line $y=x$ on reliability diagrams represents a perfectly calibrated model.}
  \label{fig:enzymes reliability}
\end{figure}

On ENZYMES dataset, we can observe our second finding that a model with relatively low test accuracy can achieve improved efficiency if the model is well-calibrated, in terms of ECE, MCE, and reliability diagram. Specifically in Fig.~\ref{fig:enzymes reliability}, in the case of the model with $\beta=1$, despite having lower model accuracy compared to other models, it results in similar or even lower inefficiency than a model with higher accuracy ($\beta=10^{-5})$ due to its low ECE, MCE, and well-calibrated behavior as observed in reliability diagram. However, it was hard to capture the first observation on ENZYMES dataset, since the GCNs with temperatures $\{10^{-5}, 10^{-4}, 10^{-3}, 10^{-2}, 10^{-1}\}$ that acheive similar accuracy did not exhibit strong calibration as visualized in Fig.~\ref{subfig:enz-b}.  

\newpage

\newpage




\end{document}